\pdfoutput=1

\documentclass[11pt]{article}

\usepackage[]{ACL2023}
\usepackage{xcolor}
\definecolor{darkgreen}{rgb}{0.0, 0.5, 0.0}
\definecolor{lightblue}{RGB}{173,216,230}
\definecolor{lightred}{RGB}{255,182,193}
\definecolor{lightgreen}{RGB}{173,255,47}
\definecolor{lightyellow}{RGB}{255,255,204}

\usepackage{amsmath,amsfonts,bm}

\def\eqref#1{equation~\ref{#1}}

\def\1{\bm{1}}

\DeclareMathAlphabet{\mathsfit}{\encodingdefault}{\sfdefault}{m}{sl}
\SetMathAlphabet{\mathsfit}{bold}{\encodingdefault}{\sfdefault}{bx}{n}

\newcommand{\highlightred}[1]{\sethlcolor{lightred}\hl{#1}}
\newcommand{\highlightyellow}[1]{\sethlcolor{lightyellow}\hl{#1}}

\usepackage{times}
\usepackage{latexsym}
\usepackage{graphicx}
\usepackage{subcaption}
\usepackage{hyperref}
\usepackage{wrapfig}
\usepackage{url}
\usepackage{graphicx}
\usepackage{multirow}
\usepackage{hyperref}
\usepackage{booktabs}
\usepackage{longtable}
\usepackage{longtable}
\usepackage{tabularx}
\usepackage{booktabs}
\usepackage{siunitx}
\usepackage{hyperref}
\usepackage{enumitem}
\usepackage{amsmath}
\usepackage{xcolor}
\usepackage{tcolorbox}
\usepackage{soul}
\usepackage{makecell}
\usepackage{multirow}
\usepackage{booktabs}
\usepackage{graphicx}
\usepackage{amsmath}
\usepackage{amssymb}
\usepackage{placeins}
\usepackage[titletoc]{appendix}

\usepackage[T1]{fontenc}

\usepackage[utf8]{inputenc}

\usepackage{microtype}

\usepackage{inconsolata}

\title{Exploring Memorization in Fine-tuned Language Models}

\author{Shenglai Zeng$^{1}$\thanks{Equal contribution.}\thanks{Work conducted during an internship at Baidu, Inc.}\,\,\,\,, Yaxin Li$^{1\ast}$, Jie Ren$^1$, Yiding Liu$^2$, Han Xu$^1$, Pengfei He$^1$ \\ \textbf{Yue Xing$^1$, Shuaiqiang Wang$^2$, Jiliang Tang$^1$, Dawei Yin$^2$ } \\ 
$^1$Michigan State University  \quad $^2$Baidu, Inc.   \\
\{zengshe1, liyaxin1, renjie3, xuhan1, hepengf1,xingyue1,  tangjili\}@msu.edu, \\
liuyiding.tanh@gmail.com, shqiang.wang@gmail.com, yindawei@acm.org
}

\begin{document}
\maketitle
\newtheorem{definition}{Definition}
\begin{abstract}
\label{abstract}


Large language models (LLMs) have shown great capabilities in various tasks but also exhibited memorization of training data, raising tremendous privacy and copyright concerns. While prior works have studied memorization during pre-training, the exploration of memorization during fine-tuning is rather limited. Compared to pre-training, fine-tuning typically involves more sensitive data and diverse objectives, thus may bring distinct privacy risks and unique memorization behaviors. In this work, we conduct the first comprehensive analysis to explore language models' (LMs) memorization during fine-tuning across tasks. Our studies with open-sourced and our own fine-tuned LMs across various tasks indicate that memorization presents a strong disparity among different fine-tuning tasks. We provide an intuitive explanation of this task disparity via sparse coding theory and unveil a strong correlation between memorization and attention score distribution. 


\end{abstract}
\vspace{-0.3cm}
\section{Introduction}
\label{Intro}
\vspace{-0.2cm}

Large language models (LLMs) have demonstrated impressive capabilities in natural language understanding and generation, enabling significant advances across diverse applications including reading comprehension, text classification, and summarization~\citep{openai2023gpt,ouyang2022training,bai2022constitutional,touvron2023llama}. However, recent works reveal that pre-trained langauge models (LMs) tend to memorize and regenerate segments of their pre-training corpus when prompted appropriately. For example,  \citet{carlini2021extracting} devised a training data extraction attack, successfully extracting hundreds of verbatim text sequences from GPT-2's training data. Existing works demonstrate that various factors can affect memorization and memorization effects grow with model scale, data duplication, and prompt length~\citep{lee2021deduplicating,kandpal2022deduplicating, carlini2022quantifying}. These findings raise privacy and confidentiality concerns, as interactions between humans and the deployed LMs could enable extraction of the memorized sensitive training data, such as phone numbers, people's names, etc. As the scale of LMs and their training data continues to expand, the privacy risks posed by memorization become increasingly serious.

In addition to pre-training, the application of LMs often involves fine-tuning on downstream tasks~\citep{touvron2023llama,chung2022scaling,ouyang2022training,longpre2023flan}, while the memorization of fine-tuning data is rather overlooked by existing studies. Compared to pre-training, fine-tuning introduces two unique perspectives with respect to memorization. \textbf{First, fine-tuning often utilizes domain-specific and private data.} For instance, developing a diagnostic chatbot \citep{yunxiang2023chatdoctor} requires collecting sensitive medical conversation data. Similarly, an academic LM \citep{beltagy2019scibert} may be trained on copyrighted essays for summarization or paraphrase generation. Leakage of such fine-tuning data can seriously violate user privacy or intellectual property rights. \textbf{Second, fine-tuning involves more complex and diverse training goals compared to pre-training. } During pre-training, the learning objective is usually language modeling from a massive unlabeled corpus (e.g., next-word prediction), which is agnostic to downstream tasks. In fine-tuning, the objective is to learn task-specific knowledge from annotated data, such as how to effectively capture the key information of a long document for summarization. The differences may induce distinct memorization behaviors and patterns during fine-tuning. Consequently, it is necessary to explore memorization for fine-tuning. Yet, it is challenging because previous insights and findings regarding the pre-trained models may not directly apply to fine-tuning. 

To bridge this gap, we focus on the memorization of LMs during fine-tuning. We study a variety of fine-tuning tasks including summarization, dialogue, question answering, machine translation, and sentiment analysis. Using an automatic plagiarism detection pipeline~\citep{lee2023language}, we examine memorization on both popular open-sourced models and the models fine-tuned for diverse tasks. In both cases, we consistently observe the existence of substantial memorization under certain tasks. Moreover, we draw several new insights and reveal potential factors that may impact the memorization of fine-tuned LMs.
Our key findings and contributions are summarized as follows: 
\begin{itemize}[noitemsep, topsep=0pt, leftmargin = *]
    \item {\textit{Disparate Memorization Across Tasks.}} Particular tasks such as summarization and dialogue present high memorization. In contrast, tasks like classification, reading comprehension, and translation exhibit low memorization. This discrepancy highlights the disparate cognitive demands these tasks need from LMs. 
    \item{\textit{Task-Dependent Scaling in Fine-tuned Memorization.}} For tasks with high memorization, the degree of memorization escalates with the increase of model size. On the other hand, for tasks with low memorization, increasing the model size does not significantly amplify the memorization.
    \item {\textit{Memorization Linked to Task Information Needs.}} We hypothesize that the varying degrees of memorization across different tasks are linked to the number of input features that LMs need to retain. We further justify this based on sparse coding models and attention patterns. Specifically,
    tasks which need to understand every detail of the input tends to memorize more from the data, and the distribution of the attention score is dense across all input-output pairs.
    
\end{itemize}

\vspace{-0.1cm}
\section{Related Work}\label{Related works}
\vspace{-0.1cm}

Powered by the transformer architecture \citep{vaswani2017attention}, in recent years, LMs such as ChatGPT \citep{ouyang2022training}, Claude \citep{bai2022constitutional}, Palm \citep{chowdhery2022palm}, Llama \citep{touvron2023llama,touvron2023llama1} and T5 \citep{raffel2020exploring} have achieved impressive performance across a wide range of natural language processing (NLP) tasks. 
These language models are pre-trained by a large amount of data to enhance their overall proficiency. Subsequently, people usually utilize various techniques \citep{chung2022scaling,ouyang2022training,houlsby2019parameter,hu2021lora,li2021prefix,lester2021power,liu2021p}  to fine-tune the pre-trained models, thus enabling them to more effectively adapt to different downstream tasks.  

The memorization behavior of pre-trained LMs has attracted increasing attention in recent years. \citet{carlini2021extracting} first proposed a data extraction attack, demonstrating that LMs tend to memorize and regenerate segments of training data.  \citet{kandpal2022deduplicating} and \citet{lee2021deduplicating} revealed that duplicated training data is more vulnerable to memorization, and de-duplication can effectively reduce memorization.  \citet{carlini2022quantifying} further quantified memorization effects, revealing that memorization grows with model scale, data duplication, and prompt length.

There are also works providing different views and understandings on memorization. For example, \citet{ippolito2022preventing} developed an efficient defense preventing memorizing the exact sentences (verbatim memorization), yet showed it fails to prevent leakage of training data. This shows the need for definitions beyond verbatim memorization. To distinguish "common"  memorization from "rare" memorization, \citet{zhang2021counterfactual} formulated a new notion of counterfactual memorization, which measures how predictions change if a particular document is deleted during training.  \citet{biderman2023emergent} investigated predictable memorization by extrapolating small or partially-trained LMs' behavior to forecast memorization in larger models. They further presented scaling laws of prediction and explored ways to improve prediction reliability.

While most literature focused on memorization during pre-training, limited work has investigated memorization in the fine-tuning stage. \citet{mireshghallah2022memorization} examined memorization risks in different fine-tuning methods for large LMs. They found that fine-tuning only the head leads to higher memorization compared to fine-tuning smaller adapter modules.  \citet{lee2023language} studied plagiarism during fine-tuning, concluding that the plagiarism patterns in fine-tuned LMs depend on corpus similarity and homogeneity. However, these studies considered fine-tuning with the same objective as pre-training, which is different from the the common practice of fine-tuning in various tasks. As a result, in this paper, we focus on the more general and realistic scenario of multifaceted fine-tuning across diverse objectives.

\vspace{-0.25cm}
\section{Preliminary}\label{Preliminary}
\vspace{-0.25cm}

In this section, we first introduce the definition of memorization and the detection methods used in this paper, and then introduce our preliminary findings on open-sourced fine-tuned LMs across various tasks.

\subsection{Definitions and Notations}

\paragraph{Definitions of memorization in literature.} In literature, there are some definitions of memorization. For example, in \citet{carlini2022quantifying}, a straightforward and strict definition is that a string $s$ is extractable with its context $p$ (with length $k$) if the concatenation $[p \| s]$ exists in the training set and $f(p)$ produces exactly the output of $s$, i.e., $f(p)=s$.  This is defined as \textbf{verbatim memorization}. In~\citet{ippolito2022preventing}, a relaxed definition of memorization is using the Bilingual Evaluation Understudy (BLEU) score. 

However, the above two definitions only consider memorizing the exact wordings of the data, instead of memorizing the meaning of the data. Therefore, in~\citet{lee2023language}, plagiarism detection tools are leveraged to to identify memorization through comparing the machine-generated text with the whole training set .  


\paragraph{Definition of fine-tuned memorization.} 
In the fine-tuning stage, models are trained for specific tasks like sentiment analysis, dialog, and summarization. We define fine-tuning as supervised training using samples \(\mathcal{D}_{\text{train}}=\left\{\left(x_i, y_i\right)\right\}_{i=1}^n\), where \(y_i\) represents the target output for input \(x_i\).
 Since the input texts usually contain more information than the output in our considered tasks, we majorly discuss the potential information leakage from the input corpus in the training set, i.e., $\mathcal{D}_\text{input} = \{x_i\}_{i=1}^n$.

To explore memorization, we follow the prompting approach~\citep{carlini2022quantifying} by dividing each $x_{i}=[p_{i} \| s_{i}]$ to a length-$k$ prefix $p_{i}$, and a suffix $s_{i}$. We further define the set of all prefixes in the training set as $P=\{p_i\}_{i=1}^n$, and the set of all suffixes as $S=\{s_i\}_{i=1}^n$. We define fine-tuned memorization as follows:

\begin{definition}[Fine-tuned memorization]
Given the fine-tuned model function \(f\), fine-tuned memorization is defined as when the model output $f(p_i)$ contains information of any $s_j\in S $, formalized by $D(f(p_i), s_j) = \text{True}$, where $D$ is a discriminative function to judge the similarity between two texts.

\end{definition}


\paragraph{Evaluation method.}  


In our practice, we input  $P=\{p_i\}_{i=1}^n$ to the model and utilize a local search engine to quickly locate suspicious texts and use a plagiarism detection tool to serve as the discriminative function $D$. In detail, given a dataset of size \( n \) with suffix space $S$, we employ the local search engine, Elasticsearch\footnote{\href{https://www.elastic.co/elasticsearch/}{https://www.elastic.co/elasticsearch/}}, to identify the top-\( K \) corpus candidates \( S_K^i = \{ s_1^i, s_2^i, \ldots, s_K^i \} \) which are similar to \( f(p_i) \). Then we utilize the PAN2014 plagiarism detection tool\footnote{\href{https://pan.webis.de/clef14/pan14-web/text-alignment.html}{https://pan.webis.de/clef14/pan14-web/text-alignment.html}} to assess the similarity between \( f(p_i) \) and each candidate \( s_j \in S_k^i \). This detection tool is capable of identifying the presence of plagiarised pairs \( (d_i, d_j) \), where \( d_i \) and \( d_j \) are sub-strings from \( f(p_i) \) and \( s_j \), respectively. The main idea of this tool is to transform text into term frequency-inverse document frequency (TF-IDF) vectors and utilize sentence similarity measures (cosine similarity) to identify plagiarism cases. We say the fine-tuned model memorizes $s_j$ if the plagiarism is confirmed. We then count the number of memorized cases and divide by $n$ to get the total memorization rate. We use $n=10000$ in our experiments. This memorization rate quantifies the memorization exhibited in the model.


 
Moreover, the detection tool can categorize memorized content into three distinct types. To provide a detailed quantification of the memorization behavior, we include these categories following the methods in prior work \citep{lee2023language}. 
\begin{itemize}[noitemsep, topsep=1pt]
    \item \textbf{Verbatim}: \( d_j \) is an exact replica of \( d_i \).
    \item \textbf{Paraphrase}\footnote{Paraphrasing is further assessed using RoBERTa and NER, classifying $p < 0.5$ as low-confidence and $p > 0.5$ as high-confidence, with both reported.}: \( d_j \) is a rephrased version of \( d_i \) 
    \item \textbf{Idea memorization}: \( d_j \) condenses  \( d_i \) into fewer sentences, or vice versa.
\end{itemize}
Generally, all these memorization types indicate that the model generates information about the suffix of training data $x$ not given as input.
More details of the detection pipeline, e.g., implementation descriptions and differences among memorization types, are included in the Appendix \ref{Details and pipeline}. Typical memorization cases are shown in Appendix \ref{Mem Cases}.


\begin{table*}[t]
\centering
\caption{Memorization rate of open-sourced LLMs fine-tuned on various tasks}
\label{tab:Open_Sourced finetune}
\resizebox{0.9\textwidth}{!}{
\begin{tabular}{@{}c|ccccccc@{}}
\toprule
Task & Dataset & \begin{tabular}[c]{@{}c@{}} Source\\Model\end{tabular} & \begin{tabular}[c]{@{}c@{}}Total\\Mem Rate \end{tabular}  & Verbatim & Idea & \begin{tabular}[c]{@{}c@{}} Paraphrase\\($p>0.5$) \end{tabular} & \begin{tabular}[c]{@{}c@{}} Paraphrase \\($p<0.5$) \end{tabular} \\
\midrule
\href{https://huggingface.co/facebook/bart-large-cnn}{Summarization} & CNN/Daily Mail  & Bart\_Large & 20.7\% & 1.3\% & 0\% & 9.8\% & 9.6\% \\
\href{https://huggingface.co/Narrativaai/BioGPT-Large-finetuned-chatdoctor}{Medical Dialog} & ChatDoctor & BioGPT & 19.6\% & 0.1\% & 3.5\% & 7.8\% & 8.2\% \\
\href{https://huggingface.co/potsawee/t5-large-generation-squad-QuestionAnswer}{Extractive QA} & SQuAD\_v2 & T5\_large & 0.1\% & 0\% & 0\% & 0\% & 0.1\% \\
\href{https://huggingface.co/potsawee/t5-large-generation-race-QuestionAnswer}{Abstractive QA} & Race & T5\_large & 0.3\% & 0\% & 0\% & 0.2\% & 0.1\% \\
\href{https://huggingface.co/facebook/wmt19-en-de}{Translation} & WMT\_19 & FSMT & 0\% & 0\% & 0\% & 0\% & 0\% \\
\href{https://huggingface.co/mrm8488/t5-base-finetuned-imdb-sentiment}{Sentiment Classification} & IMDB & T5-base & 0\% & 0\% & 0\% & 0\% & 0\% \\
\bottomrule
\end{tabular}
}
\end{table*}

\vspace{-0.3cm}
\subsection{Preliminary Findings}
\vspace{-0.2cm}

To initially explore the memorization effects during fine-tuning, we examine several popular open-sourced fine-tuned models from HuggingFace that were fine-tuned on 6 representative tasks.We attach the descriptions of all models and datasets used in Appendix \ref{Datasets and model}.  These tasks include summarization, medical dialog, question and answering (QA), translation, and sentiment analysis. 
The preliminary results in Table~\ref{tab:Open_Sourced finetune} suggest that substantial memorization of the fine-tuning data occurs in fine-tuning. For summarization and medical dialog models, we identified total memorization rate of 20.7\% and 19.6\%, respectively. These high rates could imply potential privacy violations or copyright issues.
Furthermore, the level of memorization varies across tasks.
Models fine-tuned for summarization and medical dialog exhibit high memorization, while models for remaining tasks show much lower memorization. These observations motivate further in-depth analysis to validate the observed task-specific memorization behavior in Section~\ref{Ex1} and the possible scaling effect in Section~\ref{Ex2} to fine-tune models by ourselves. Furthermore, in Section \ref{Ex3}, we provide an in-depth analysis and understanding of the potential reason behind the observed disparate fine-tuned memorization across tasks.


\vspace{-0.15cm}
\section{Disparate Memorization Across Tasks}
\label{Ex1}
\vspace{-0.15cm}

Our preliminary study demonstrates that the memorization of fine-tuned models varies on different fine-tuning tasks. However, the causes of such difference are still unclear, which could be fine-tuning datasets or model architectures.
To clarify this, in Section~\ref{sec:41}, we 
control the impact of different variables to more precisely explore the relation between fine-tuning task and the memorization effect. In Section \ref{sec:impact factor}, we further examine the impact of decoding methods and prefix length.  Note that in our fine-tuning process, we ensure that our fine-tuned models have satisfactory performance on downstream tasks, see Appendix \ref{finetune performance}. 



\begin{table*}[htbp]
\centering
\caption{Memorization rate of T5-base fine-tuned on various tasks}
\label{tab:self finetune}
\resizebox{0.8\textwidth}{!}{
\begin{tabular}{@{}c|c|c|ccccc@{}}
\toprule
Task & Dataset & Model & \begin{tabular}[c]{@{}c@{}} Total \\Mem Rate \end{tabular} & Verbatim & Idea & \begin{tabular}[c]{@{}c@{}} Paraphrase \\ (P>0.5) \end{tabular} & \begin{tabular}[c]{@{}c@{}} Paraphrase \\ (P<0.5) \end{tabular} \\
\midrule
\multirow{3}{*}{Summarization} & \multirow{3}{*}{Multi-news} & T5-base & 13.98\% & 3.58\% & 0.66\% & 4.28\% & 5.46\% \\
 & & T5-finetuned & 22.33\% & 4.23\% & 0.65\% & 6.23\% & 11.22\% \\
 & & Difference & $\uparrow$8.35\% & $\uparrow$0.65\% & $\downarrow$0.01\% & $\uparrow$1.95\% & $\uparrow$5.76\% \\
\midrule
\multirow{3}{*}{Dialog} & \multirow{3}{*}{chatdoctor} & T5-base & 1.60\% & 0.03\% & 1.04\% & 0.11\% & 0.42\% \\
 & & T5-finetuned & 8.27\% & 0.02\% & 1.41\% & 1.75\% & 5.09\% \\
 & & Difference & $\uparrow$6.67\% & $\downarrow$0.01\% & $\uparrow$0.37\% & $\uparrow$1.64\% & $\uparrow$4.67\% \\
\midrule
\multirow{3}{*}{\begin{tabular}[c]{@{}c@{}} Sentiment \\ Classification \end{tabular}} & \multirow{3}{*}{imdb} & T5-base & 0.78\% & 0.05\% & 0.37\% & 0.16\% & 0.20\% \\
 & & T5-finetuned & 0.80\% & 0.04\% & 0.30\% & 0.17\% & 0.29\% \\
 & & Difference & $\uparrow$0.02\% & $\downarrow$0.01\% & $\downarrow$0.07\% & $\uparrow$0.01\% & $\uparrow$0.09\% \\
\midrule
\multirow{3}{*}{\begin{tabular}[c]{@{}c@{}} Reading \\ Comprehension \end{tabular}} & \multirow{3}{*}{Squad\_v2} & T5-base & 0.08\% & 0.02\% & 0.00\% & 0.01\% & 0.05\% \\
 & & T5-finetuned & 0.15\% & 0.04\% & 0.00\% & 0.05\% & 0.06\% \\
 & & Difference & $\uparrow$0.07\% & $\uparrow$0.02\% & - & $\uparrow$0.04\% & $\uparrow$0.01\% \\
\midrule
\multirow{3}{*}{Translation} & \multirow{3}{*}{wmt} & T5-base & 0.00\% & 0.00\% & 0.00\% & 0.00\% & 0.00\% \\
 & & T5-finetuned & 0.00\% & 0.00\% & 0.00\% & 0.00\% & 0.00\% \\
 & & Difference & - & - & - & - & - \\
\bottomrule
\end{tabular}}
\end{table*}

\vspace{-0.2cm}

\subsection{Fine-tuned with Fixed Pre-trained LM and Dataset}\label{sec:41}

\vspace{-0.1cm}

\paragraph{Fine-tuned on T5-base LM.}  To eliminate the potential impact from the pre-trained LM, we conduct experiments to fine-tune the same pre-trained T5-base model\footnote{We also fine-tuned GPT-Neo models and found consistent findings with T5, we report the results in Appendix \ref{gptneo-ft} and Table \ref{tab:self fine-tune gptneo}.} for different fine-tuning tasks. We treat all the tasks as generative tasks (like instruction-tuning) and do not add any additional modules (e.g., MLP). Details of the datasets and memorization results are presented in Table~\ref{tab:self finetune}\footnote{We provide the statistical significance testing of memorization rate in Appendix \ref{Statistic Test}. }. In case a part of the fine-tuning data appears in the pre-training data, we report the memorization rate of the pre-trained model\footnote{In the context of pre-trained models, we continue to utilize the prefixes from the fine-tuning dataset for evaluating memorization. This process is detailed in Appendix \ref{Pipeline}.} and present the change in the memorization rate before and after fine-tuning. Table~\ref{tab:self finetune} clearly shows a substantial total memorization rate  and memorization gain for summarization tasks (22.3\%, $\uparrow$8.35\%) and medical dialogue (8.27\%, $\uparrow$6.67\%). Meanwhile, the memorization and the gain in fine-tuning are much lower for reading comprehension (0.15\%, $\uparrow$0.07\%), translation (0.0\%, $\uparrow$0.0\%), and sentiment classification (0.8\%, $\uparrow$0.02\%). These observations are consistent with our preliminary findings on open-sourced models and suggest that fine-tuned memorization with the same pre-trained LM architecture still demonstrates a strong task disparity. 


\paragraph{Fine-tuned on different tasks with the same dataset.}

\begin{figure}[ht]
  \centering
  \includegraphics[width=0.35\textwidth]{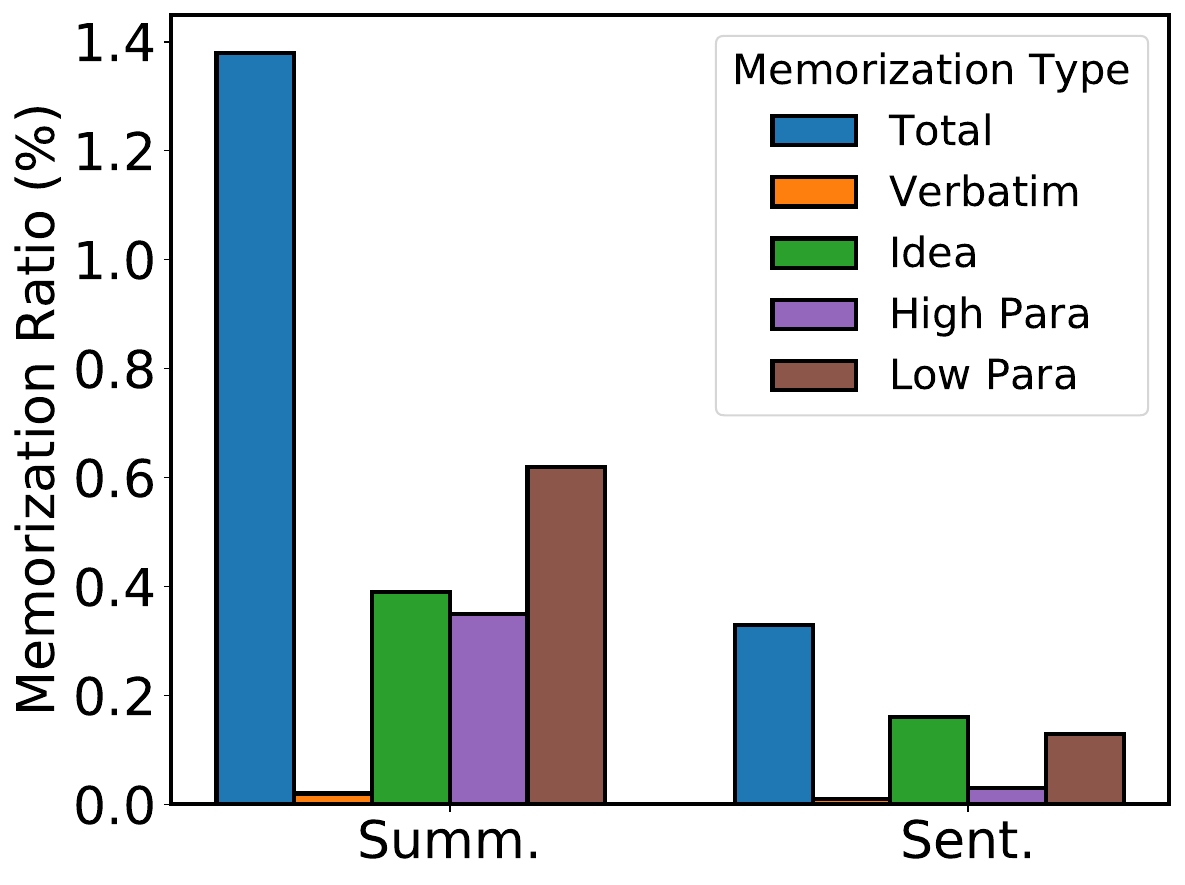}
  \caption{\small Memorization of T5-base fine-tuned on RentTheRunway.}
  \label{fig:Rent}
\end{figure}

In this experiment, we investigate the memorization of different tasks fine-tuned on the same pre-trained LM with the same dataset. Specifically, we fine-tune the T5-base model on RentTheRunway dataset \citep{misra2018decomposing}, which contains self-reported clothing fit feedback from customers along with additional metadata. Each product has multiple attributes, including customer reviews, ratings, review summaries, review dates, etc. 
It allows us to fine-tune the same pre-trained model (i.e., T5-base), with the identical inputs (i.e., the customer reviews), for different task objectives (i.e., review summarization and sentiment classification). 
Note that we map the ratings into positive and negative labels for fine-tuning sentiment classification model. 
The memorization performance of these two models is shown in Figure~\ref{fig:Rent}. The results demonstrate that the summarization model exhibits higher memorization compared to the classification model, which validates that the task objective impacts the memorization of fine-tuned models.

\vspace{-0.4cm}

\subsection{Further Probing}
\vspace{-0.2cm}
\label{sec:impact factor}
\begin{figure*}[t]
\centering
\resizebox{0.9\textwidth}{!}{%
    \begin{minipage}{\textwidth}
        \subfloat[ Prefix length]{\includegraphics[width=.49\textwidth]{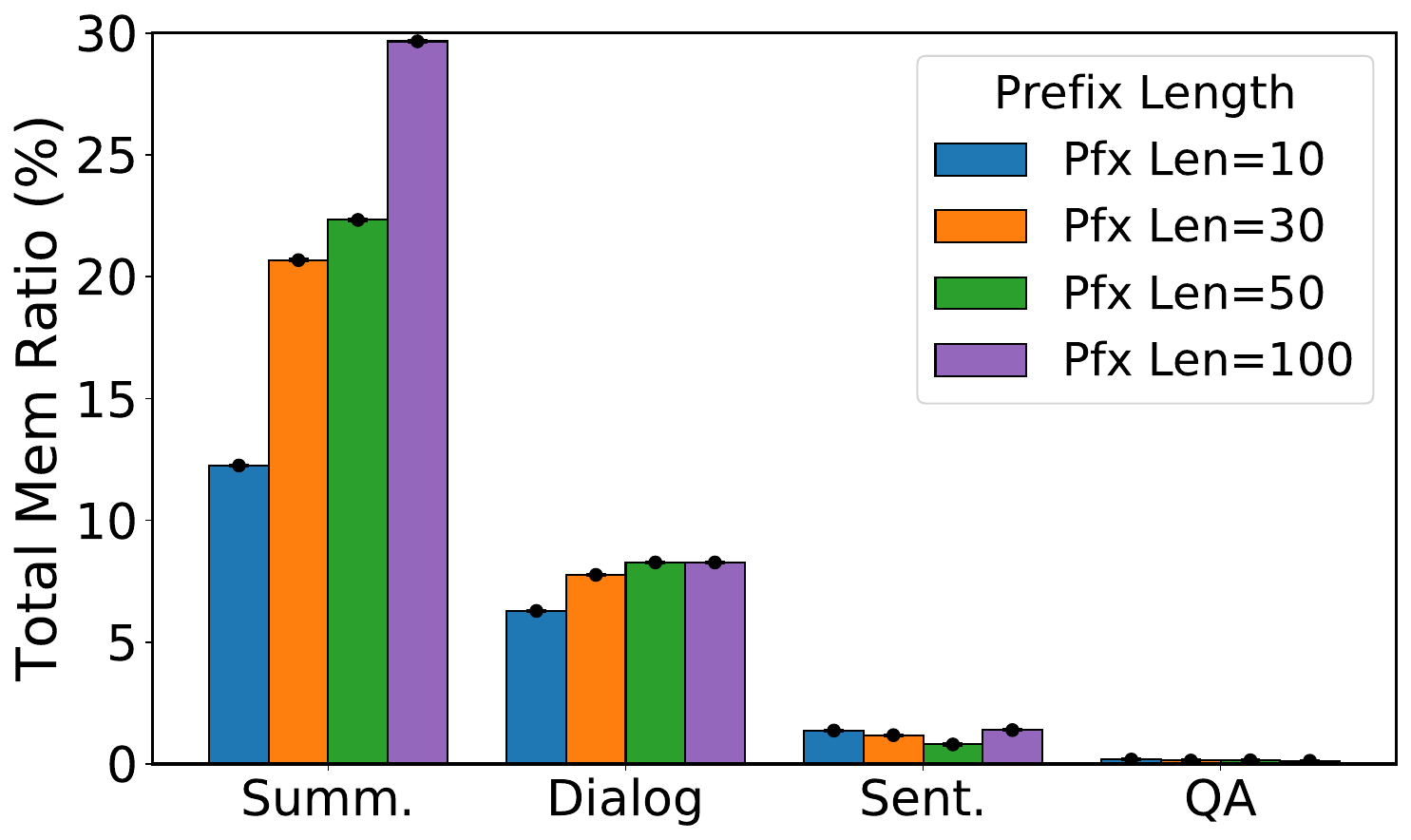}
        \label{fig:Ablation_prefix}}
        \subfloat[ Sampling methods]{\includegraphics[width=.49\textwidth]{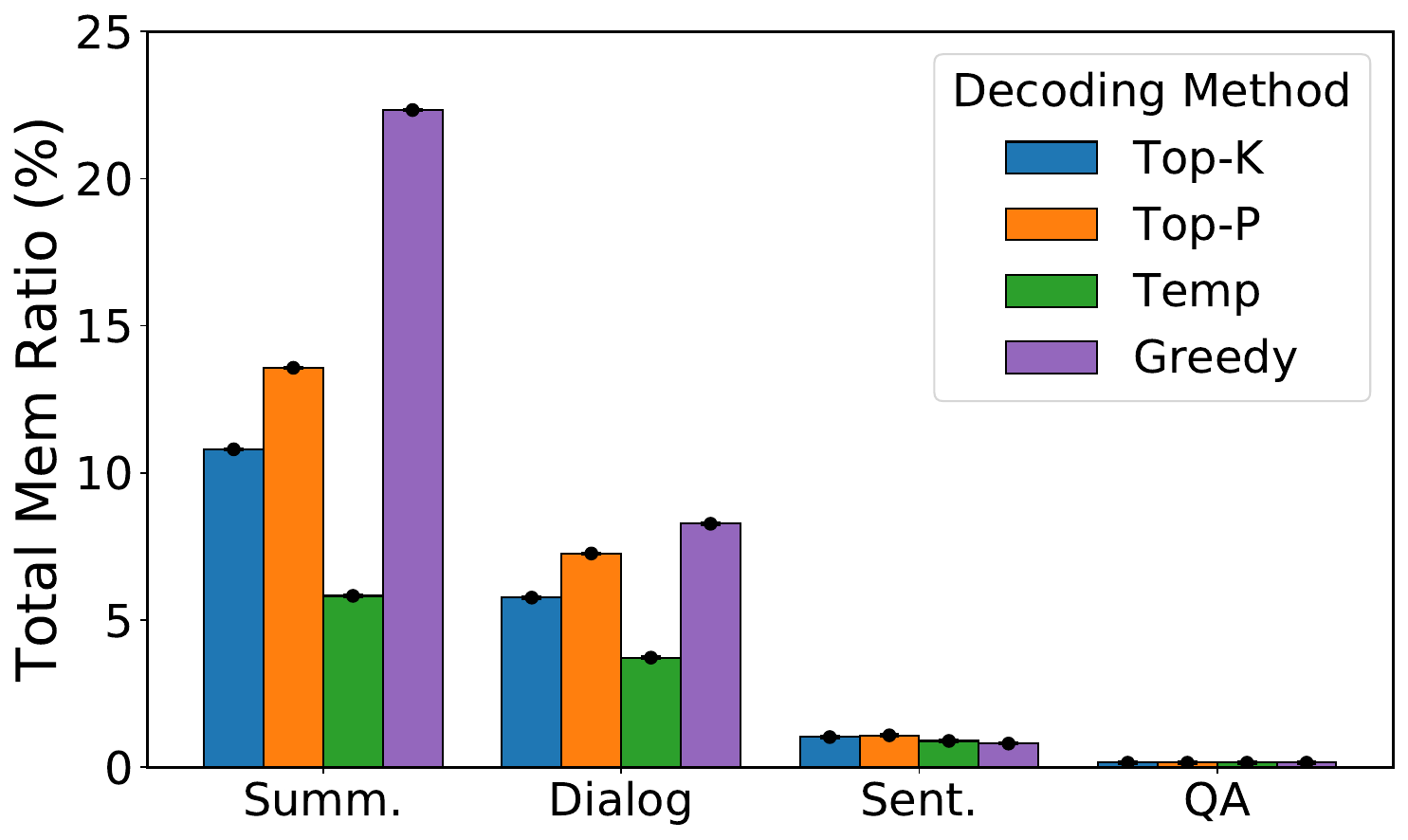}
        \label{fig:Ablation_decoding}}
    \end{minipage}
}

\caption{Impact of prefix length and sampling methods on memorization }
\label{fig:Ablation}
\end{figure*}

The memorization behavior of pre-trained models has been shown to be influenced by factors such as generation sampling methods and input prefix lengths \citep{carlini2022quantifying, lee2023language}. In this subsection, we extend our investigation to explore how these factors affect memorization in fine-tuned models. We also conduct an ablation study on the impact of training epochs on memorization and report the results in Appendix \ref{ablation_training} and Table~\ref{tab: ablation_epochs}. 

\paragraph{Input prefix length.} We vary the length of the prefix of inputs and report the main results in Figure \ref{fig:Ablation_prefix}. 
 We find that the length of prefix tokens influences memorization differently across tasks. In summarization and dialogue tasks, the total memorization rate tends to increase with longer prefixes, aligning with existing research on pre-trained memorization. However, in sentiment classification and QA, altering the prefix length does not significantly affect memorization. More results in Appendix \ref{prefix_ab} and Table \ref{tab: t5_prefix_length}. \textbf{Despite these variations, a consistent disparity in memorization across different tasks persists, regardless of the prefix length.}
\vspace{-0.15cm}
 \paragraph{Generation sampling.}
We study the impact of different generation sampling methods including (i) top-k (k=40) sampling, (ii) top-p (p=0.8) sampling and (iii) changing the temperature to T=1, and report the main results in Figure \ref{fig:Ablation_decoding} and more results in Appendix \ref{decoding_ab} and Table \ref{tab: t5_decoding}. It is observed that sampling affects memorization differently across tasks: it lowers memorization in high-memorization tasks like summarization and dialogue, but has a negligible or even increasing effect on memorization in low-memorization tasks such as sentiment classification. \textbf{Despite these variations, a significant, consistent disparity in memorization remains across different tasks, indicating an intrinsic, task-specific inclination towards memorization that is not significantly altered by sampling methods.}

\vspace{-0.1cm}
\section{Memorization Scaling Behavior of Fine-tuned LMs}
\label{Ex2}
\vspace{-0.1cm}

It is evident that memorization in pre-trained models tends to increase with model size. To understand the scaling behavior of memorization in fine-tuned models, we conduct a systematic analysis on fine-tuned models, comparing the memorization in various tasks using different sizes of the T5 model: T5-small (60M), T5-base (220M), T5-large (770M), and T5-xl (3B). 

\begin{figure*}[t]
\centering
\resizebox{\textwidth}{!}{%
    \begin{minipage}{0.9\textwidth}
        \subfloat[Summarization]{\includegraphics[width=.25\textwidth]{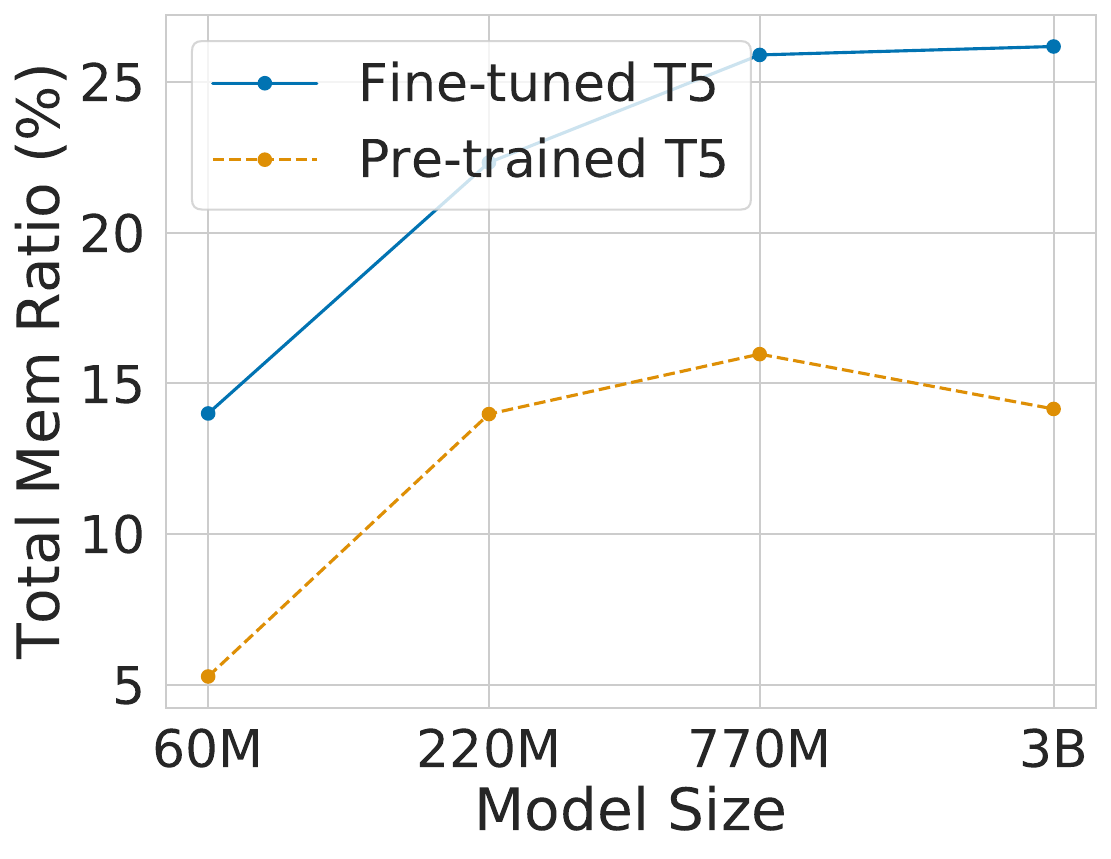}
        \label{fig:summary-scaling}}
        \subfloat[Dialog]{\includegraphics[width=.25\textwidth]{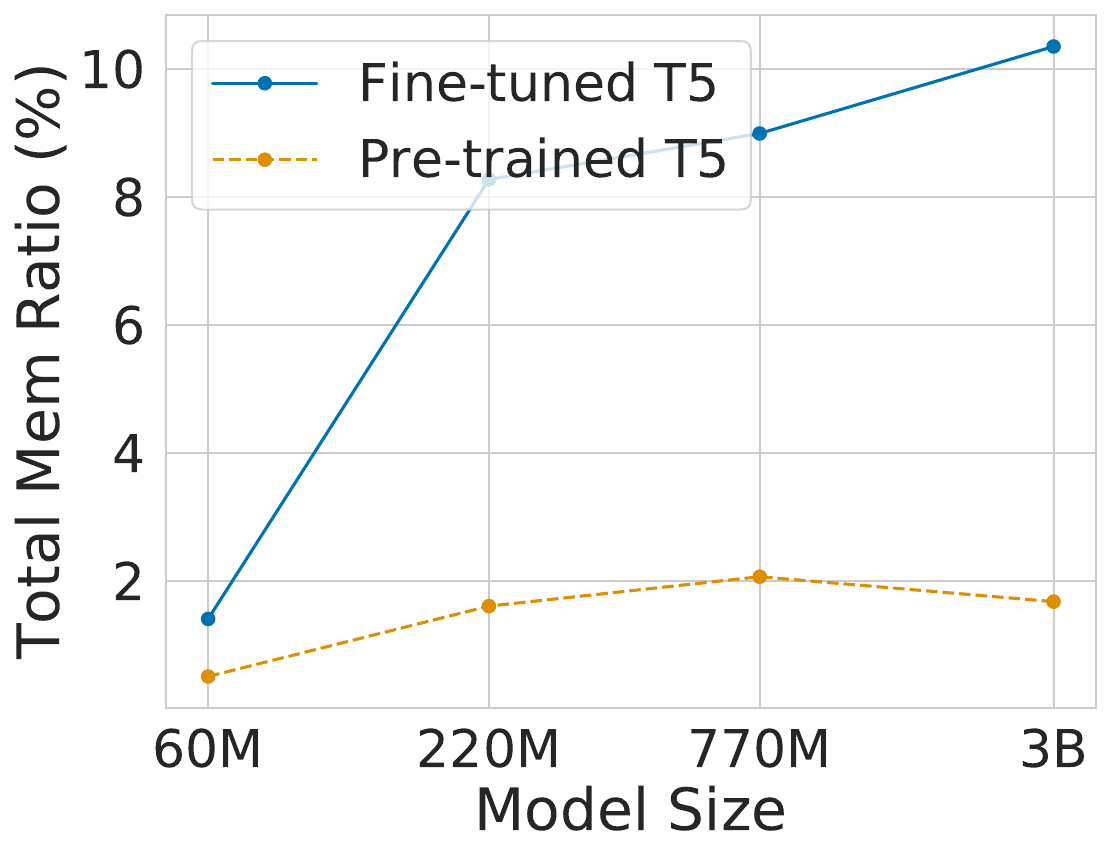}
        \label{fig:dialog-scaling}}
        \subfloat[Sentiment classification]{\includegraphics[width=.25\textwidth]{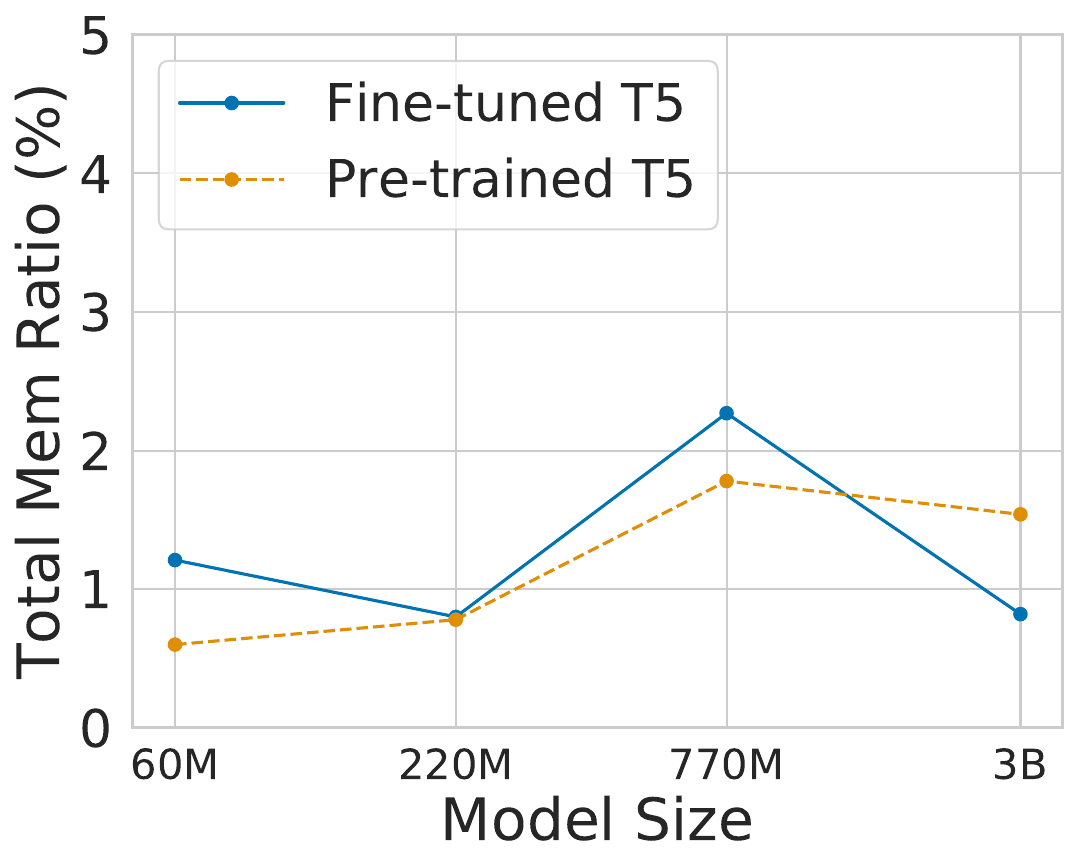}
        \label{fig:sentiment-scaling}}
        \subfloat[Extractive QA]{\includegraphics[width=.25\textwidth]{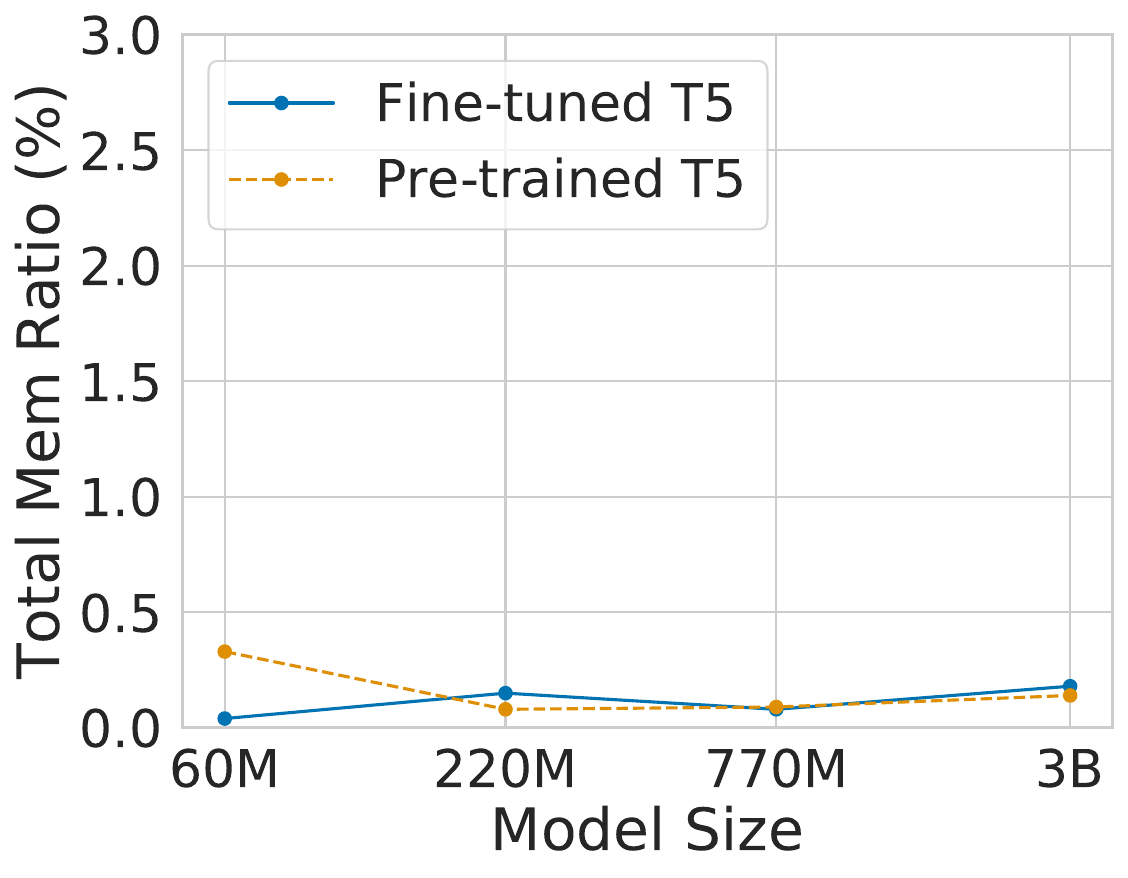}
        \label{fig:QA-scaling}}
    \end{minipage}
}

\caption{Scaling behavior of fine-tuned memorization}
\label{fig:Scaling}
\end{figure*}
\vspace{-0.2cm}
\subsection{High-memorization Tasks }
\vspace{-0.2cm}

We explore the scaling behavior in high-memorization tasks, specifically focusing on summarization (fine-tuned on Multi-news) and dialogue (fine-tuned on Chatdoctor). The results can be found in Figures \ref{fig:summary-scaling} and \ref{fig:dialog-scaling} respectively. We can see that memorization in fine-tuning increases with model size. As a benchmark, we also provide the memorization rate for the pre-trained model, for which we can see that when increasing from a model size of 220M, the memorization rate does not further increase much. These two observations in the fine-tuned model and the pre-trained model together reveal that the fine-tuned model is memorizing information from the fine-tuning data, indicating severe privacy threats when scaling up the models in these tasks.

\vspace{-0.35cm}
\subsection{Low-memorization Tasks }
\vspace{-0.3cm}
In contrast to high-memorization tasks, low-memorization tasks such as sentiment classification and question answering exhibit different scaling behaviors. As illustrated in Figures \ref{fig:sentiment-scaling} and \ref{fig:QA-scaling}, an increase in model size does not result in a rise in the memorization rate, and the memorization rate is consistently low. This suggests that even when large models are fine-tuned on these tasks, the possibility of memorization and outputting fine-tuning data is relatively low.
 \vspace{-0.5cm}
\section{Understanding the Memorization Disparity}
\label{Ex3}

\begin{figure*}[t]
\centering
\resizebox{0.95\textwidth}{!}{%
    \begin{minipage}{\textwidth}
        \subfloat[Avg-Summ.]{\includegraphics[width=.25\textwidth]{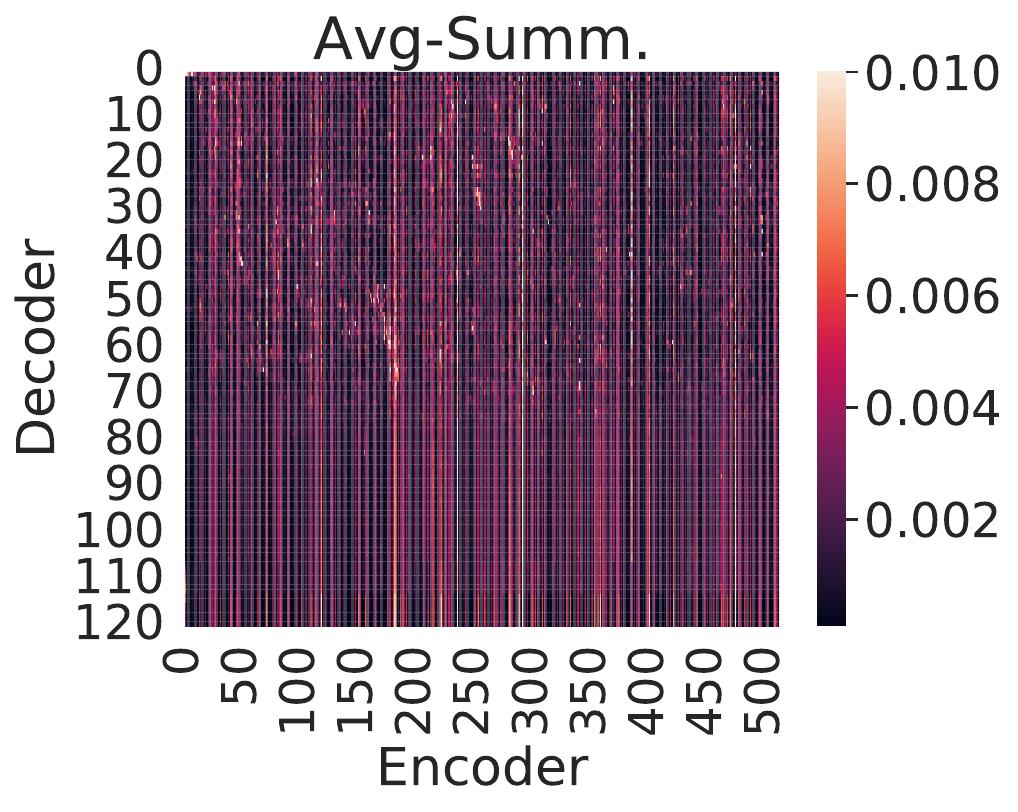}
        \label{fig:summary-heatmap}}
        \subfloat[Sample-Summ.]{\includegraphics[width=.25\textwidth]{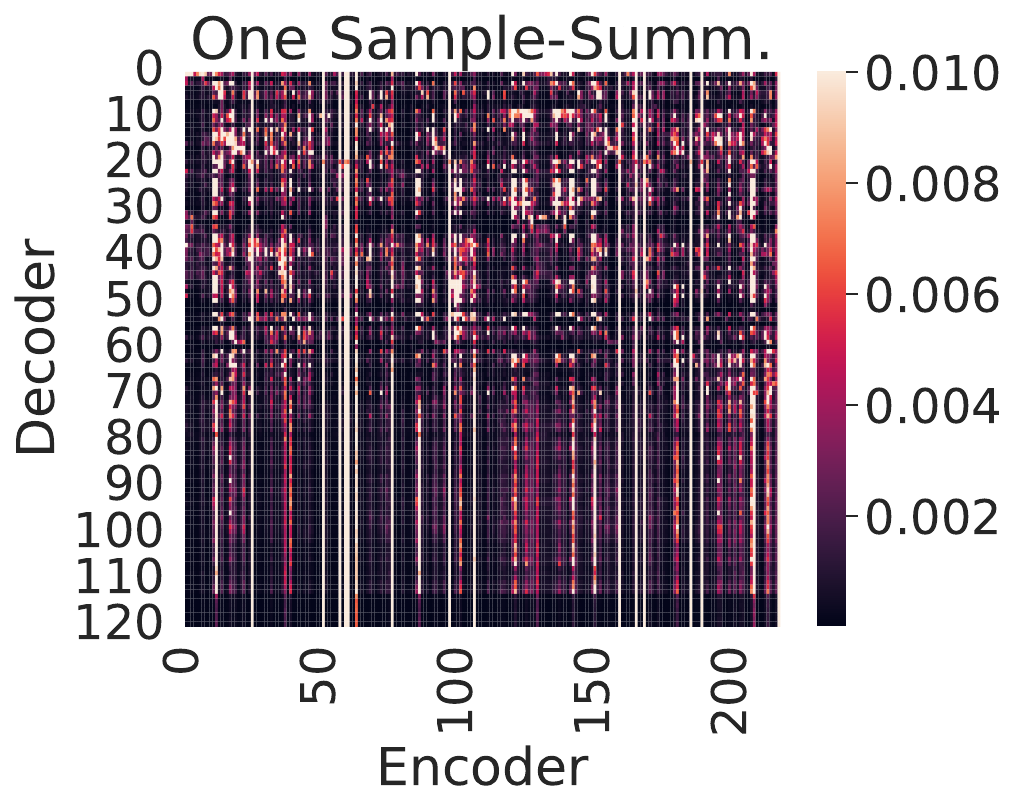}
        \label{fig:summary-A}}
        \subfloat[Avg-Dialog]{\includegraphics[width=.25\textwidth]{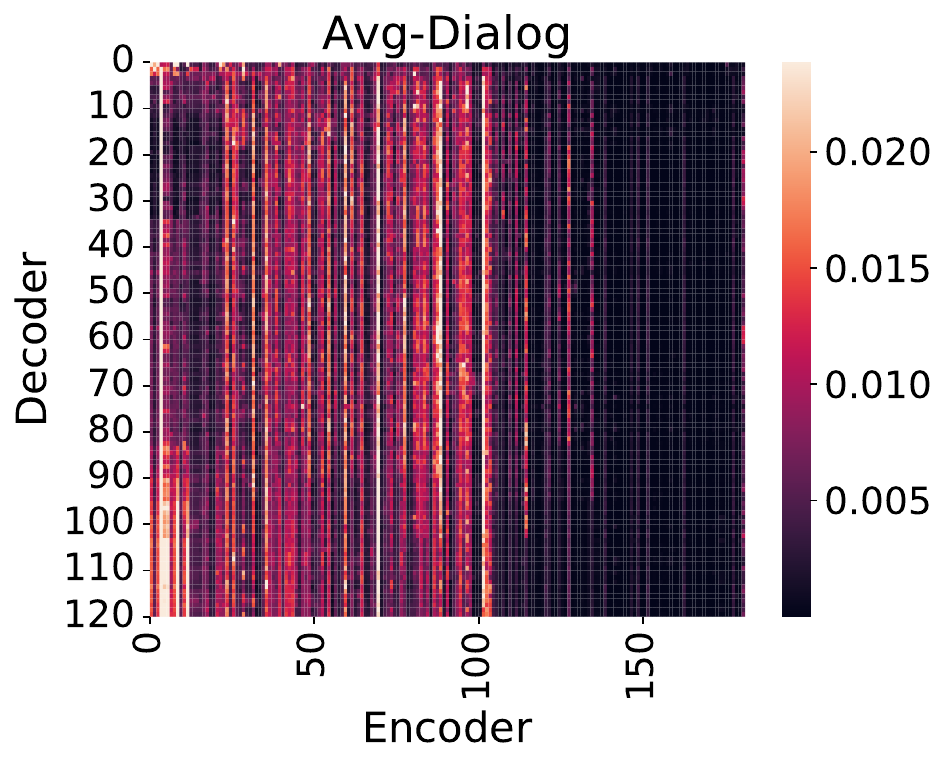}
        \label{fig:dialog-heatmap}}
        \subfloat[Sample-Dialog]{\includegraphics[width=.25\textwidth]{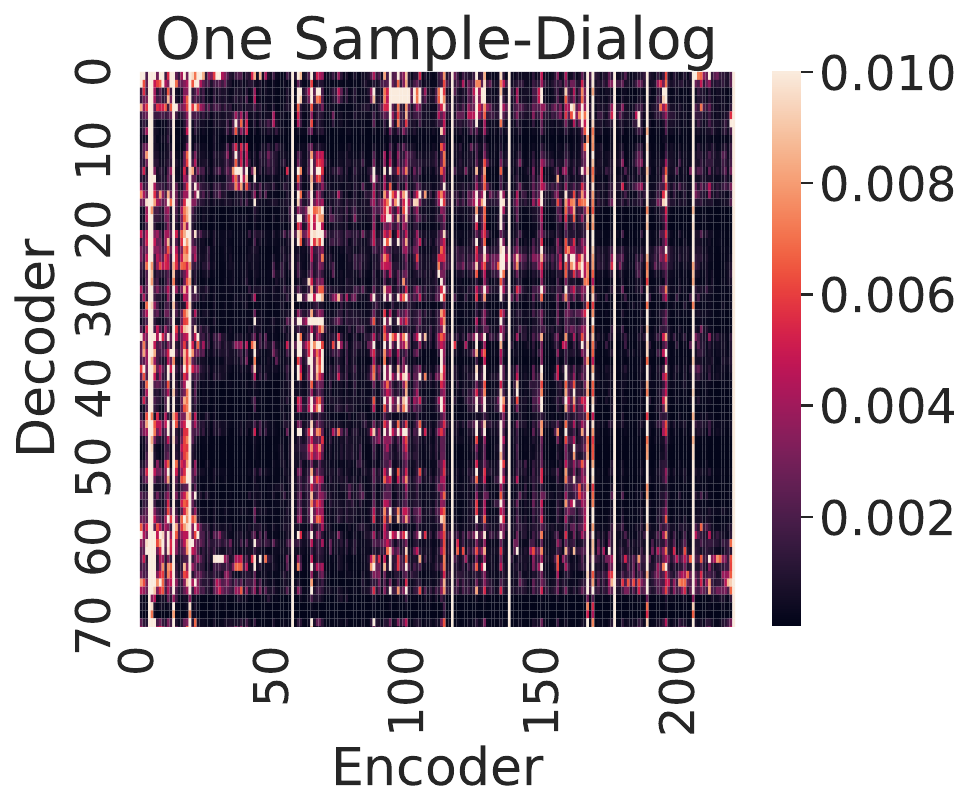}
        \label{fig:dialog-A}}\\
        \subfloat[Avg-Sent.]{\includegraphics[width=.25\textwidth]{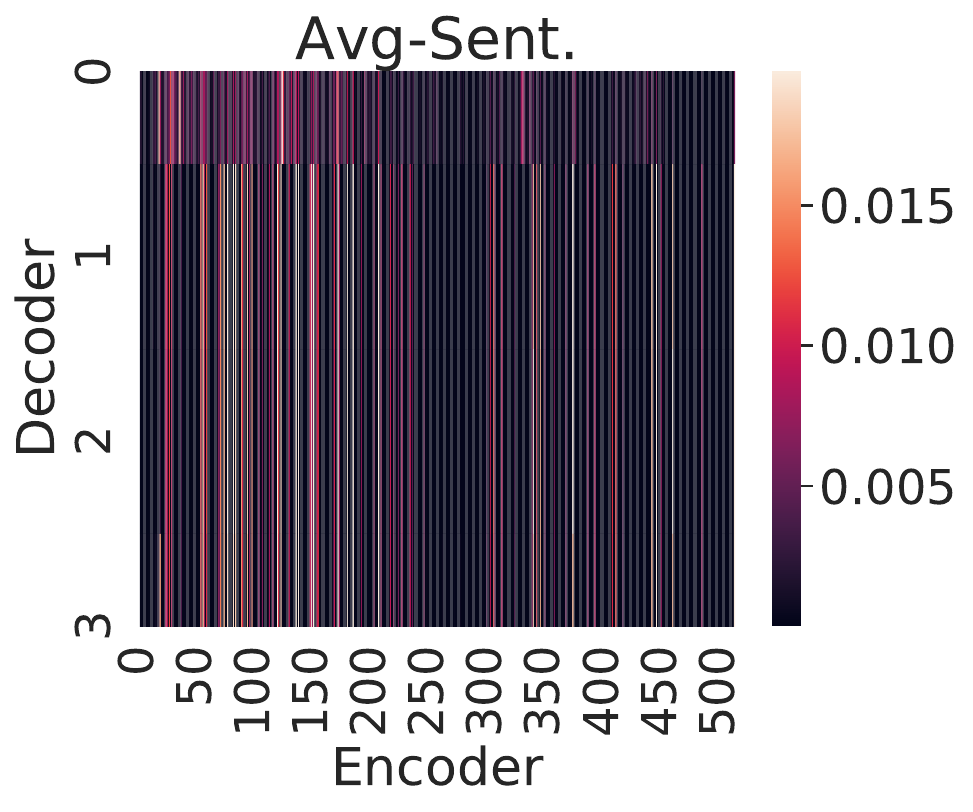}
        \label{fig:sentiment-heatmap}}
        \subfloat[Sample-Sent.]
        {\includegraphics[width=.25\textwidth]{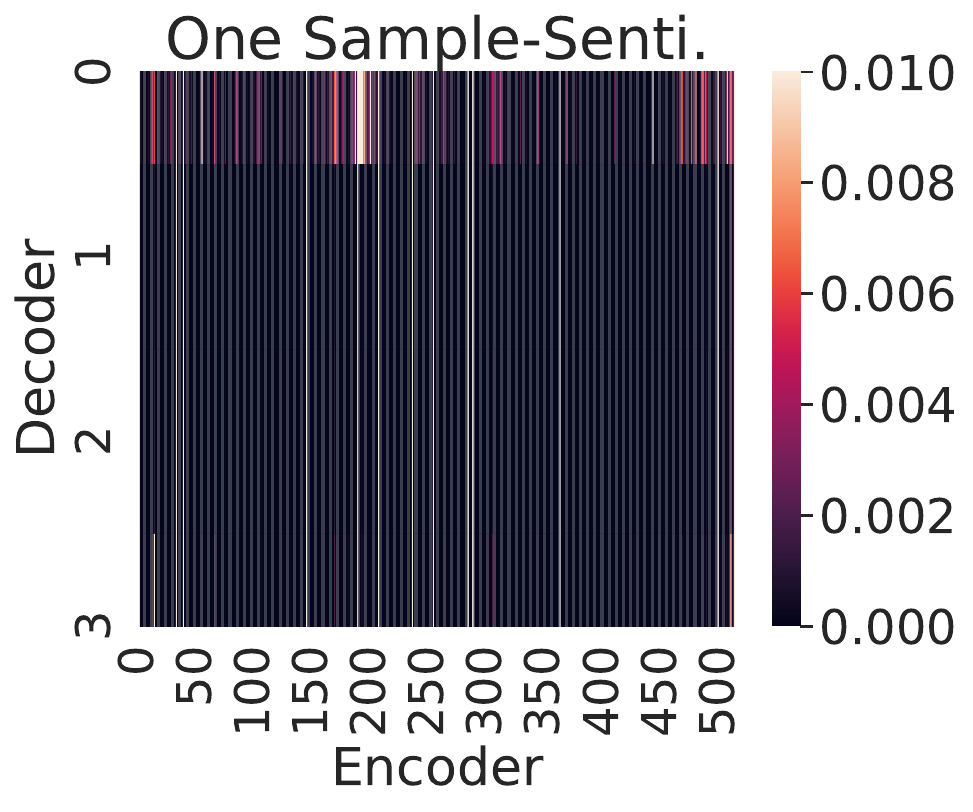}
        \label{fig:sentiment-A}}
        \subfloat[Avg-QA]{\includegraphics[width=.25\textwidth]{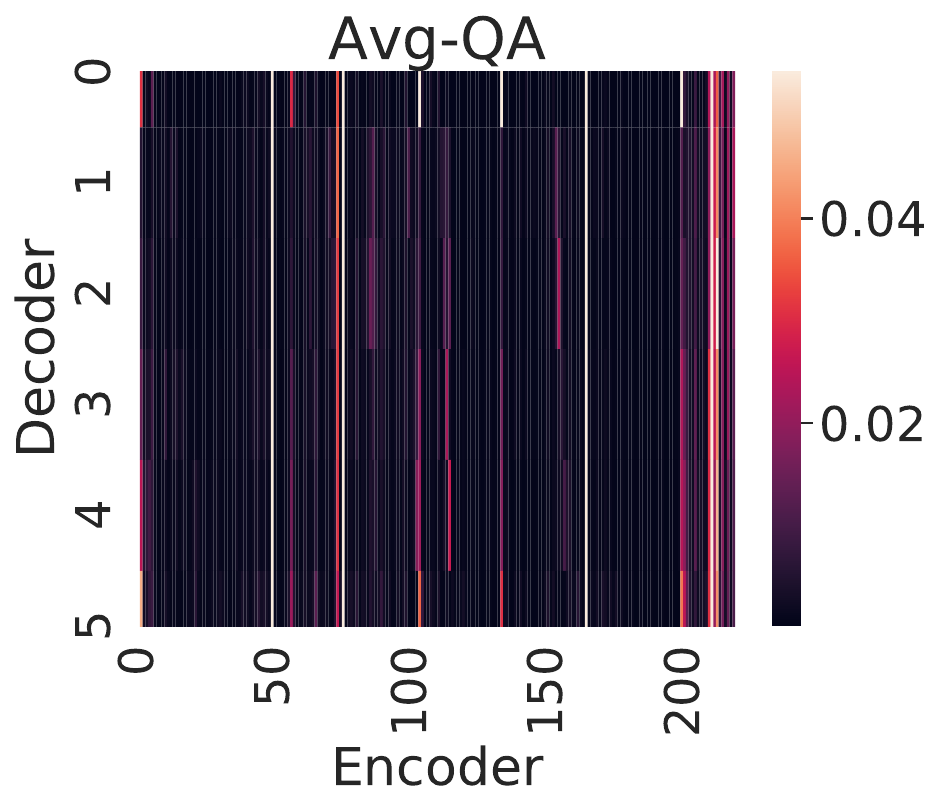}
        \label{fig:QA-heatmap}}
        \subfloat[Sample-QA]{\includegraphics[width=.25\textwidth]{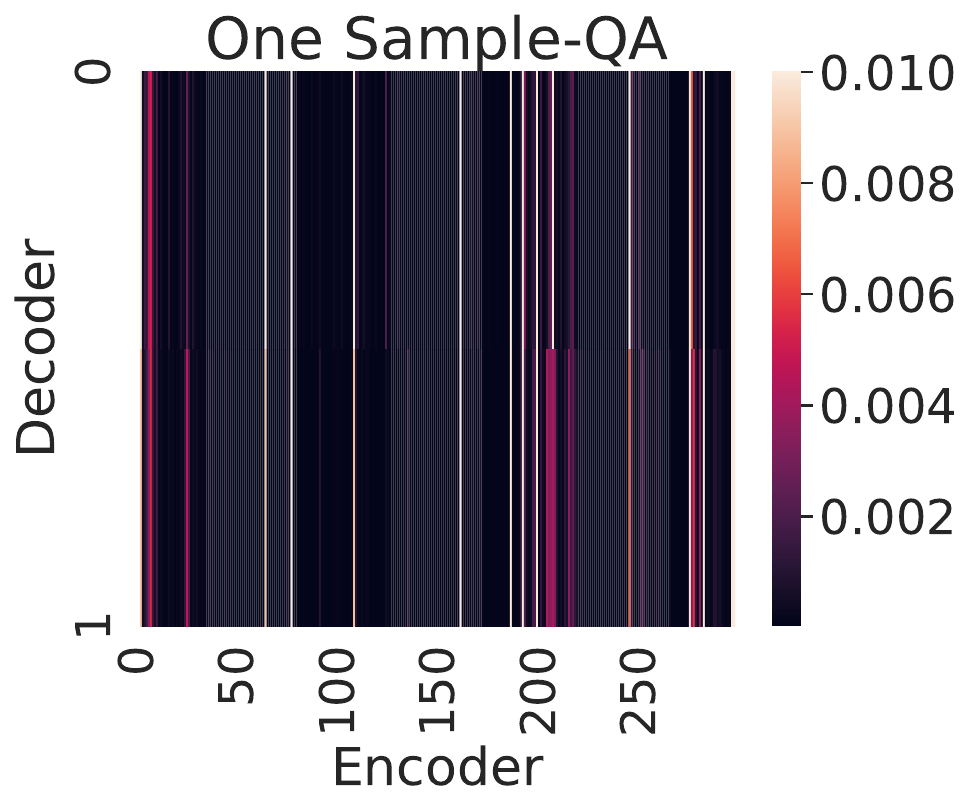}
        \label{fig:QA-A}}
        
    \end{minipage}
}

\caption{Decoder-encoder attention heatmaps for (a, b) Summarization, (c, d) Dialog, (e, f) Sentiment Analysis, and (g, h) QA. (a, c, e, g) show average heatmaps from 10 samples, while (b, d, f, h) show heatmaps from a single sample.}
\label{fig:heatmap}
\end{figure*}

\vspace{-0.2cm}
In the previous sections, we empirically examine the memorization rate and scaling behavior among a variety of tasks, and demonstrate the discrepancy between high- and low-memorization tasks. In this section, we provide understanding and evidence on the underlying reason behind such disparity of fine-tuned memorization.

\vspace{-0.2cm}
\subsection{Correlation between Memorization and Task-specific Information}\label{sec:sparse_coding}

In this subsection, we aim to investigate the question: \textit{why do different fine-tuning tasks present different memorization behaviors?} We conjecture that the memorization behavior might be closely related to the information needed to fulfill certain language tasks. Intuitively, for language tasks such as sentiment analysis or extractive QA, only a few words or sentences are enough for the model to complete the task. For example, one can determine the sentiment based on some specific words in the sentiment, and can answer a question based on certain pieces of information. In this case, the model only needs to learn specific key features and is less likely to memorize the other data. On the other hand, for tasks such as summarization and dialog, they require the model to learn more input features to complete the task, as the essential information from these inputs is also reflected in the output. As a result, the fine-tuning process will encode more input knowledge from the data in the model parameters, leading to potential concerns of memorization. In the following, we provide a conceptual discussion based on the sparse decoding model. The sparse coding model is a method that represents original data by focusing on its key features, using only the most crucial elements to efficiently express the core information, which is a popular model for modeling text and vision data.~\citep{arora2015latent,arora2018linear,olshausen1997sparse,olshausen2004sparse} 
\vspace{-0.2cm}
\paragraph{Sparse coding model.} 
Denote an observed text data as $ Z \in \mathbb{R}^{d\times D}$ where $D$ and $d$ represent the sequence length and the length of the embedding respectively.
The basic assumption of sparse coding is that the data $Z$ comes from combinations of a few hidden features. 
We use $X$ to represent these hidden features and consider the following relation: 
\begin{equation}
  Z=UXV
    \label{eq:Z=UXV}
\end{equation}
where $X \in R^{k\times K}$, $U\in\mathbb{R}^{d\times k}$, $V\in\mathbb{R}^{K\times D}$ ,  $k \leq d$, $K \leq D$. Each column of $U$ is a unit vector and orthogonal with each other. Each row of $V$ is a unit vector and orthogonal with each other. Given the above formulation, each element in $Z$ is a linear combination of the elements from $X$. Compared to previous literature, our assumption is a simplified version of the original sparse coding model as we do not further impose noise in the data generation model. Besides, we also modify the original 1D feature $X$ in sparse coding into 2D for our model. 
\vspace{-0.2cm}
\paragraph{Task complexity.} 

Under the sparse coding model, we further assume that the output can be fully expressed by a linear transformation of $X$. However, different fine-tuning tasks may differ in \textit{how much input information is needed by the task}. We present two perspectives below to illustrate why "complex tasks" may have more memorization.

First, the number of parameters to connect $Z$ with the final target is related to the task-specific information. Consider we have a \textit{``simple task''} (e.g., sentiment analysis) where the model output is only one scalar (preference) decided by some certain features or a combination of $X$. In the simplest case, the scalar is a linear function of $X$, $a^{\top}Xb$, where $a\in\mathbb{R}^k$ and $b\in\mathbb{R}^K$. In this case, the loss function for a sentiment classification model  $f_{cls}$ can be defined as:
\begin{equation}
\label{cls_loss}
    l(f_{cls}(Z),a^{\top}Xb).
\end{equation}
If we further assume the loss functions as square loss, the best solution of $f_{cls}$ is:
   $f_{cls}(Z)=(a')^{\top}Zb'$
where $a^{\top}U^{\top}:=a'$ and $V^{\top}b:=b'$. 
It means that the model only needs to learn two vector parameters $a'$ and $b'$.
On the other hand, for tasks such as summarization, the output text is desired to contain all key information from the input $Z$. We consider the following loss for the summarization task $f_{sum}$:
\begin{equation}\label{eq:complex}
    l(f_{sum}(Z), X)
.\end{equation}
In the above formulation, the output $f_{sum}(Z)$ contains all information about $X$. We denote such a task as a \textit{``complex task''}. With the squared loss, the best solution of $f_{sum}$ is $f_{sum}(Z)=U^{\top}Z V^{\top}$. Comparing the above two tasks, for simple tasks like classification, the model just requires a small amount of information to learn $a',b'$ ($d+D$), while for complex tasks such as summarization, the model needs to learn ($dk+DK$)  parameters of $U'=U^{\top},V'=V^{\top}$.
Further, the model which learns more information from the data tends to memorize more. The expression in Equation~\ref{eq:complex} makes model inversion attack possible. As the model learns  $U', V'$, the attacker can conduct an attack via $Z=U'^{T}XV'^{T}$, which means one can use $f(Z)$ to recover the input data $Z$. 

Second,  the sparsity of the learned matrix ($U$,$V$) or vectors ($a'$,$b'$) may also vary, indicating different amounts of information needed and leading to different complexity of the task. For example, in sentiment classification, what the network actually learns depends on the sparsity of $b'$.  If $b'$ is sparse, it means that we can simply pick several words from the sequence and determine the class.

\vspace{-0.2cm}
\subsection{Attention Distributions}
\vspace{-0.2cm}
{While the memorization disparity is possibly related to the task-specific information, the attention distribution in the transformer may also capture the contribution of each token's information to completing the task. In this section, we study whether the attention scores can be viewed as an indicator of the memorization ability of the task. 

For the fine-tuned models in Table \ref{tab:self finetune}, we generate their attention score heatmaps\footnote{We present the heatmap of the translation task in Appendix\ref{attention_translation}.}. Figure \ref{fig:heatmap} shows the distribution of the attention scores of the last decoder-encoder attention block in each model. Different layers of decoding-encoding attention scores are also visualized in Appendix ~\ref{attention_layers}\footnote{Across various layers, consistent patterns emerge in the encoder-decoder attention mechanism. Owing to this uniformity, we primarily report on the final layer in Figure \ref{fig:heatmap}, which is closest to the output.}. Our focus on encoder-decoder attention layers is their ability to capture the information across input features for each output.
We also theoretically discuss the correlation between attention score maps and task information needs in Appendix~\ref{Analyse}.


In Figure \ref{fig:heatmap}, the horizontal axis represents input tokens, and the vertical axis indicates output tokens. The brightness represents the averaged multi-head decoder-encoder attention scores between input-output token pairs. Each horizontal line shows the attention score distribution of an output token across input tokens. We visualize the attention heatmaps of a single data sample and the average attention heatmaps of 10 random samples, padded to the longest length of the batch and truncated to a maximum of 512 tokens. More attention maps of different samples are visualized in Appendix \ref{sample attention}.

The heatmaps shown in Figure \ref{fig:heatmap} reveal clear differences in attention patterns among tasks. For high-memorization tasks (i.e., summarization and dialog), attention scores are more evenly distributed across input tokens. In contrast, for low-memorization tasks (e.g., sentiment classification and extractive QA), the attention is concentrated on a few positions while almost zero for other positions. The observed patterns suggest that the information needed to successfully complete each task varies. The attention score distribution for summarization and dialog implies that models must extract every detail from the input, increasing the possibility of memorization. Concentrated scores for sentiment classification and extractive QA indicate that only key information is required, reducing the tendency to memorize the fine-tuning data. 


In Appendix \ref{attention_t5base}, we present the attention scores of the T5-base model. Our findings also align with the fine-tuned model, and we observe a pattern of high memorization-intensive attention for complex tasks and concentrated attention for simpler ones. 
This aligns with our intuition that the attention pattern is a fundamental characteristic of the task. 
%
To utilize the attention score as a tool to predict memorization, prior to fine-tuning a model for a specific task, developers can assess attention patterns. The assessment can be done using the pre-trained model. It helps predict the memorization during fine-tuning.

\vspace{-0.3cm}
\section{Conclusions}
\label{Conclusion}
\vspace{-0.3cm}
In this paper, we conduct extensive experiments to investigate the memorization behavior of fine-tuned LMs among various tasks. Utilizing an automatic detection pipeline, we are able to evaluate the memorization in numerous tasks and datasets. In addition, we provide understandings of the memorization disparity among tasks based on a sparse coding theory. Our analysis reveals a strong correlation between attention scores and memorization.  
\section{Limitations}
Our study primarily utilizes white-box models from the T5 family for evaluation, such as T5-small, T5-base, T5-large, and T5-XL. Black-box models like ChatGPT were not included due to our inability to fine-tune them directly. Additionally, there is scope to expand the variety of datasets and tasks in future research. 

Theoretically, we employ sparse coding theory to articulate our hypothesis that the observed memorization differences across tasks stem from their varying informational requirements ("True features" in sparse coding theory). We draw on recent theoretical developments\citep{zhang2023trained, deora2023optimization,wu2023many} that apply sparse coding in NLP and leverage these theories to support our reasoning. Nevertheless, fully extending this theory to precisely account for the complexities of non-linear or large-scale models remains an unresolved challenge in the theoretical community.

While our study addresses the memorization behavior of models fine-tuned on single tasks, the investigation into models fine-tuned on multiple tasks is still unexplored and presents an opportunity for future research.

\section{Ethics Statement}
The phenomenon of fine-tuned memorization in language models has notable social implications. Firstly, it raises privacy concerns, especially in tasks where sensitive information, like personal dialogue, is involved. The ability of these models to retain and potentially disclose private data necessitates corresponding data protection measures. Secondly, from a utility perspective, while memorization enhances performance in certain tasks, it also underscores the need for balancing accuracy with responsible data handling in AI systems.
\bibliography{anthology}

\clearpage

\appendix


\section{Theoretical Analyze}
\label{Analyse}
To understand why attention scores can be used as an indicator for memorization, we provide some theoretical intuition on the relation between attention scores and information needed for the task. We first use the classification task to explain the relation between attention scores and information density. Then we extend the intuition to discuss complex tasks.
\paragraph{Attention score and memorization in classification} 

We still consider the sparse coding model mentioned in Eq.\ref{eq:Z=UXV}, and continue to use the notations of Section \ref{sec:sparse_coding}. We use the classification task for simplicity.  As mentioned, for classification tasks, the best solution of Eq.\ref{cls_loss} is $f_{cls}(Z)=(a')^{\top}Zb'$, where $a^{\top}U^{\top}:=a'$ and $V^{\top}b:=b'$.
In classification, whether a task is more complex or not depends on the sparsity of $b'$, and we justify that the\textbf{ sparsity of $b'$ directly affects the attention score pattern. } We then mathematically define the neural network architecture. To ease the derivation, we consider
\begin{equation}
    f(Z) = W^VZ \cdot\text{softmax}\left( (W^KZ)^{\top} (W^QZ)\right),
\end{equation}
with $W^V, W^K, W^Q$ all in $\mathbb{R}^{d\times d}$. The softmax operation is conducted column-wise. Since the output $f(Z)$ is a matrix rather than a scalar, for the classification task, we further multiply two vectors on the two sides of $f(Z)$ to get output scalar $y'$, i.e., $v_1^{\top}f(Z)v_2\in\mathbb{R}$, and $v_1$ and $v_2$ can be either trainable or arbitrary.  As mentioned in Section \ref{sec:sparse_coding}, the target of the classification task can be represented as $y=a^{\top}Xb$. The loss term of Eq.\ref{cls_loss} can then be written as: 
\begin{equation}
    l(v_1^{\top}f(Z)v_2,a^{\top}Xb).
    \label{eq:l_f_cls_2}
\end{equation}
Aligning the neural network output $v_1^{\top}f(Z)v_2$ with $a^{\top}Xb$, it is easy to see that to better reduce the loss value, we need $\text{softmax}\left( (W^KZ)^{\top} (W^QZ)\right)v_2\in\mathbb{R}^D$ better aligned with $b'$. As a result, when measuring the effect of the input $Z$ on $v_1^{\top}f(Z)v_2\in\mathbb{R}$, e.g., Figure \ref{fig:sentiment-heatmap}, the weighted pattern $\text{softmax}\left( (W^KZ)^{\top} (W^QZ)\right)v_2$  has a similar sparsity as $b'$. Recall that $b'$ is the task-specific vector the model needs to learn, thus the analysis above suggests that \textbf{the attention score\footnote{Note that the attention matrix $\text{softmax}\left( (W^KZ)^{\top} (W^QZ)\right)$ itself is $\mathbb{R}^{D\times D}$ and is not aggregated for the output value, and $\text{softmax}\left( (W^KZ)^{\top} (W^QZ)\right)v_2$ is the final aggregated attention.} has a similar sparsity pattern as the sparsity of the information the model needs to learn. }
\paragraph{Complex tasks}

For more complex tasks, the simplistic single-layer single-head attention analysis as the above is not enough to handle it,  and we need to use a larger architecture. Intuitively, with more features to learn in the task, the architecture will be more likely to memorize each feature comprehensively. We identify two key drivers of this behavior. First, each output token relies on information distributed across multiple input tokens. As shown in Figure \ref{fig:summary-heatmap}, each row has multiple high attention scores across different input tokens. Second, each output token often exhibits selectivity for a different subset of input tokens, leading to divergence in attention distributions across rows in Figure \ref{fig:summary-heatmap}. To conclude, these two factors may result in dense heatmap patterns compared to the concentrated heatmaps of simpler tasks.

\section{Details of Evaluation Pipeline}
\label{Details and pipeline}
\subsection{Evaluation Process}
\label{Pipeline}
In this section, we provide a detailed overview of the evaluation tools and methodologies employed. The evaluation framework comprises three distinct processes: prompting, searching, and detection. The prompting method has been extensively utilized in prior research on pre-trained memorization\citep{carlini2022quantifying}, and the search and detection methods, based on Elasticsearch and PAN-2014 detection tools, were previously adopted by \citet{lee2023language}.  

\paragraph{Prompting.} In the evaluation phase, the input data \( x_{i} \) is segmented into two parts: a prefix \( p_{i} \) and a suffix \( s_{i} \). We input the prefixes \( \{p_i\}_{i=1}^n \) into the model without any task-specific instructions to obtain \( f(p_i) \). For our experiments, we select \( n=10,000 \) samples from each dataset. The standard prefix length \( k \) is set to 50 tokens. However, for tasks with input sentences shorter than 50 tokens, such as translation tasks, a reduced prefix length of 15 tokens is used. The testing procedure is consistent across both base and fine-tuned models. This involves inputting the prefix of the fine-tuning data into the models and comparing the suffixes. Notably, for the base models, the prefix of the fine-tuning data is still used in the test, regardless of whether the model has been fine-tuned on that specific dataset or not. This approach helps determine if the model retains the data, suggesting its presence in the pre-training set.


\paragraph{Searching.}In the search phase of our experiment, we employ Elasticsearch, a distributed, RESTful search and analytics engine based on the open-source Lucene library. Elasticsearch leverages the Okapi-BM25 algorithm, a widely-used bag-of-words ranking function, allowing for efficient storage, searching, and near real-time analysis of large data volumes. We upload all suffixes \( \{s_i\}_{i=1}^n \) into Elasticsearch and use the set \( \{f(p_i)\}_{i=1}^n \) as our query documents. For our analysis, we set \( K = 10 \), indicating that only the top-10 most relevant candidates for each query are retrieved for subsequent memorization detection.


\paragraph{Detection.} After we get suspicious sentence pairs $f(p_i)$ and $s_j$, we input them to a publicly available PAN2014 plagiarism detection tool $D$ to see if $D(f(p_i),s_j)= \text{True}$. In general, the detection tool will detect the presence of plagiarised word piece pairs \( (d_i, d_j) \) a, where \( d_i \) and \( d_j \) are word pieces from \( f(p_i) \) and \( s_j \), respectively and then compare $(d_i,d_j)$ to identify the category of memorization.  Here we set the minimal match threshold as at least 50 characters(approximately 15 tokens).

The detailed process includes (1) preprocessing text; (2) identifying obfuscation types; (3) seeding to find candidate pairs via sentence similarity; (4) extension by clustering similar fragments; and (5) filtering out overlaps. They transform sentences into TF-IDF vectors and calculate similarity using dice and cosine measures, with adaptive parameters selected by testing on the obfuscation corpus.  Here we set the minimal match threshold as at least 50 characters(approximately 20 tokens). We also utilize additional validation steps after retrieving paraphrased text segments as \cite{sanchez2015adaptive}. The post-processing involves chunking segments into sentences using NLTK's tokenizer, then applying a RoBERTa-based paraphrase identification model and Named Entity Recognition (NER) on the sentences. Specifically, we check sentence pairs - if any pair has a paraphrase detection probability score between 0.5 and 0.99, we accept it as high-confidence paraphrasing, otherwise, we identify it as low-confidence paraphrasing.
\subsection{Memorization Types}

\paragraph{Difference between memorization types.}
Here we will distinguish 3 types of memorization. First, verbatim memorization means exact copies of words or phrases without transformation. In the cases of paraphrase and idea memorization, the output is not identical to the original text but shares similar meanings. While paraphrase plagiarism focuses on sentence-to-sentence transformation, idea plagiarism involves summarizing the key points of a larger text segment into a more condensed form (or expanding it). In Table~\ref{Memorization Types}, we give a simple example to differentiate the difference between 3 types of memorization conceptually.

In practice of the PAN2014-detection, It starts by identifying closely matched short document fragments (referred to as 'seeds') and then expands these seeds into longer text segments. This is achieved by clustering these fragments based on their separation and employing a 'maxgap' threshold parameter to form coherent clusters.They experimentally find out the most suitable threshold for different plagiarism datasets so that those parameters could be used for the detection of a specific type of memorization. In other words, each memorization case will be \textbf{counted only once} and there will not be overlapping across different categories.

\paragraph{Distinguishing idea memorization from summarization.}
It's important to differentiate idea memorization—\textit{condensing key points of a larger text segment}—from summarization tasks. Note that in our approach, only the prefix (initial tokens) of a text is input to the model. Summarization means the model summarizing this prefix without revealing the remaining suffix of the text. In contrast, idea memorization involves the model generating information about the suffix. In our experiments, we assess similarity by comparing the generated text $f(p)$ with the suffix $s$, rather than with the prefix $p$ or the entire text $x$. In table \ref{Idea vs summ}, we use a simple example to illustrate the difference.

\begin{table*}[ht]
\caption{Difference between 3 Memorization Types}
\label{Memorization Types}
\centering
\begin{tabular}{p{15cm}}

\hline
\textbf{Examples of 3 Memorization Types } \\
\hline
\textbf{Verbatim:}\\
Text A: My name is Jack\\
Text B: My name is Jack\\

\textbf{Paraphrase:} \\
Text A: My name is Jack\\
Text B: Jack is my name \\

\textbf{Idea plagiarism:}\\
Text A: A boy tell me in the class that his name is Jack\\
Text B: A boy is Jack\\
\hline
\end{tabular}

\label{tab: type difference}
\end{table*}

\begin{table*}[ht]
\caption{Difference between Idea Memorization and Summarization}
\label{Idea vs summ}
\centering
\begin{tabular}{p{15cm}}

\hline
\textbf{Idea Memorization vs Summarization } \\
\hline
\textbf{Training data:} I am not comfortable from the beginning of the month, I am 20 years old, height 51, height 51, weight 40kg. I have been pregnant for 6 months and I can not stop vomiting.
 \\

\textbf{Input:}  I am not comfortable from the begining of the month,I am 20 years\\
\textbf{Output:}  20 years old, height 5 1 \& weight 40kg(Memorization)\\
\textbf{Output:} 20 years old woman feels bad for a month.(Summarization)\\
\# Note: We only compare output with \textit{"height 51, height 51, weight 40kg. I have been pregnant for 6 months and I can not stop vomiting."} to identify memorization.\\
\hline
\end{tabular}

\label{tab:idea vs mem}
\end{table*}

\begin{table*}[htbp]
\centering
\caption{Memorization of fine-tuned\_T5 with various epochs.}
\label{tab: ablation_epochs}
\resizebox{0.5\textwidth}{!}{%
\begin{tabular}{c|c|c|c}
\toprule
Task & Dataset & Epochs & \begin{tabular}[c]{@{}c@{}} Total\\Mem Rate \end{tabular} \\ 
\midrule
\multirow{4}{*}{Dialog} & \multirow{4}{*}{HealthCareMagic} & 1 & 1.60\% \\
 &  & 3 & 3.30\% \\
 &  & 5 & 6.22\% \\
 &  & 10 & 8.27\% \\
\midrule
\multirow{4}{*}{Sentiment} & \multirow{4}{*}{IMDB} & 1 & 0.78\% \\
 &  & 3 & 0.79\% \\
 &  & 5 & 0.80\% \\
 &  & 10 & 0.79\% \\
\midrule
\multirow{4}{*}{Summarization} & \multirow{4}{*}{Multi\_news} & 1 & 14.12\% \\
 &  & 3 & 14.32\% \\
 &  & 5 & 22.33\% \\
 &  & 10 & 22.32\% \\
\midrule
\multirow{4}{*}{QA} & \multirow{4}{*}{Squad\_v2} & 1 & 0.08\% \\
 &  & 3 & 0.12\% \\
 &  & 5 & 0.10\% \\
 &  & 10 & 0.15\% \\
\bottomrule
\end{tabular}
}
\end{table*}

\begin{table*}[htbp]
\centering
\caption{Memorization of fine-tuned\_T5 with various sampling methods.}
\label{tab: t5_decoding}
\resizebox{\textwidth}{!}{%
\begin{tabular}{c|c|c|ccccc}
\toprule
Task & Dataset & Decoding& \begin{tabular}[c]{@{}c@{}} Total\\Mem Rate \end{tabular} & Verbatim  & Idea  & \begin{tabular}[c]{@{}c@{}}Paraphrase\\ ($P<0.5$)\end{tabular}  & \begin{tabular}[c]{@{}c@{}}Paraphrase\\ ($P>0.5$)\end{tabular}  \\ 
\midrule
\multirow{4}{*}{Dialog} & \multirow{4}{*}{HealthCareMagic} & Top-K & 5.76\% & 0.05\% & 0.38\% & 0.90\% & 4.43\% \\
 &  & Top-p & 7.26\% & 0.06\% & 0.48\% & 1.35\% & 5.37\% \\
 &  & Temp & 3.72\% & 0.02\% & 0.18\% & 0.58\% & 2.94\% \\
 &  & Greedy & 8.27\% & 0.02\% & 1.41\% & 1.75\% & 5.09\% \\
\midrule
\multirow{4}{*}{Sentiment} & \multirow{4}{*}{IMDB} & Top-K & 1.02\% & 0.01\% & 0.13\% & 0.18\% & 0.70\% \\
 &  & Top-p & 1.08\% & 0.01\% & 0.12\% & 0.22\% & 0.73\% \\
 &  & Temp & 0.89\% & 0.01\% & 0.07\% & 0.19\% & 0.62\% \\
 &  & Greedy & 0.80\% & 0.04\% & 0.30\% & 0.17\% & 0.29\% \\
\midrule
\multirow{4}{*}{Summarization} & \multirow{4}{*}{Multi\_news} & Top-K & 10.80\% & 2.54\% & 0.34\% & 1.94\% & 5.98\% \\
 &  & Top-p & 13.57\% & 4.07\% & 0.54\% & 2.26\% & 6.70\% \\
 &  & Temp & 5.82\% & 1.28\% & 0.23\% & 0.83\% & 3.48\% \\
 &  & Greedy & 22.33\% & 4.23\% & 0.65\% & 6.23\% & 11.22\% \\
 \midrule
 \multirow{4}{*}{QA} & \multirow{4}{*}{Squad\_v2} & Top-K & 0.15\% & 0.04\% & 0.00\% & 0.05\% & 0.06\% \\
 &  & Top-p & 0.15\% & 0.04\% & 0.00\% & 0.05\% & 0.06\% \\
 &  & Temp & 0.15\% & 0.04\% & 0.00\% & 0.05\% & 0.06\% \\
 &  & Gready & 0.15\% & 0.04\% & 0.00\% & 0.05\% & 0.06\% \\
\bottomrule
\end{tabular}
}
\end{table*}

\begin{table*}[htbp]
\centering
\caption{ Memorization of fine-tuned T5 with varying prefix lengths.}
\label{tab: t5_prefix_length}
\resizebox{\textwidth}{!}{%
\begin{tabular}{c|c|c|ccccc}
\toprule
Task & Dataset & \begin{tabular}[c]{@{}c@{}} Prefix\\length \end{tabular}& \begin{tabular}[c]{@{}c@{}} Total\\Mem Rate \end{tabular}& Verbatim & Idea & \begin{tabular}[c]{@{}c@{}} Paraphrase\\($p>0.5$) \end{tabular} & \begin{tabular}[c]{@{}c@{}} Paraphrase\\($p<0.5$)\end{tabular} \\ 
\midrule
\multirow{4}{*}{Summarization} & \multirow{4}{*}{Multi\_news} & 10 & 12.25\% & 1.74\% & 2.85\% & 0.88\% & 6.78\% \\
 &  & 30 & 20.68\% & 7.07\% & 1.41\% & 3.05\% & 9.15\% \\
 &  & 50 & 22.33\% & 4.23\% & 0.65\% & 6.23\% & 11.22\% \\
 &  & 100 & 29.66\% & 10.61\% & 0.79\% & 4.27\% & 13.99\% \\
\midrule
\multirow{3}{*}{Dialog} & \multirow{3}{*}{HealthCareMagic} & 10 & 6.28\% & 0.03\% & 1.94\% & 0.85\% & 3.46\% \\
 &  & 30 & 7.76\% & 0.04\% & 1.28\% & 1.72\% & 4.72\% \\
 &  & 50 & 8.27\% & 0.02\% & 1.41\% & 1.75\% & 5.09\% \\
\midrule
\multirow{4}{*}{Sentiment} & \multirow{4}{*}{IMDB} & 10 & 1.37\% & 0.00\% & 1.12\% & 0.06\% & 0.19\% \\
 &  & 30 & 1.18\% & 0.01\% & 0.51\% & 0.15\% & 0.51\% \\
 &  & 50 & 0.80\% & 0.04\% & 0.30\% & 0.17\% & 0.29\% \\
 &  & 100 & 1.39\% & 0.05\% & 0.23\% & 0.33\% & 0.78\% \\
\midrule
\multirow{4}{*}{QA} & \multirow{4}{*}{Squad\_v2} & 10 & 0.18\% & 0.05\% & 0.00\% & 0.06\% & 0.07\% \\
 &  & 30 & 0.14\% & 0.04\% & 0.00\% & 0.04\% & 0.06\% \\
 &  & 50 & 0.15\% & 0.04\% & 0.00\% & 0.05\% & 0.06\% \\
 &  & 100 & 0.13\% & 0.03\% & 0.01\% & 0.04\% & 0.05\% \\

\bottomrule
\end{tabular}
}
\end{table*}

\begin{table*}[ht]

\caption{Examples of training data from different tasks}
\label{training template}
\centering
\begin{tabular}{p{15cm}}

\hline
\textbf{Summarization} \\
\hline
\textbf{Prompt:} Please summarize the following paragraph: \\
\textbf{Input:}\\
...A fresh update on the U.S. employment situation for January hits the wires at 8:30 a.m. New York time offering one of the most important snapshots on how the economy fared during the previous month. Expectations are for 203,000 new jobs to be created, according to economists polled by Dow Jones Newswires, compared to 227,000 jobs added in February. The unemployment rate is expected to hold steady at 8.3\%.  ...\\
\textbf{Output:}\\
...The unemployment rate dropped to 8.2\% last month, but the economy only added 120,000 jobs, when 203,000 new jobs had been predicted, according to today's jobs report. \\
\textbf{Training Format:}\\
Training input = \textbf{Prompt} + \textbf{Input}, Training label =  \textbf{Output} \\
\hline
\textbf{Sentiment classification} \\
\hline
\textbf{Prompt:} \\
Please classify the sentiment of the following paragraph: \\
\textbf{Input:} \\
"Foxes" is a serious look at the consequences of growing up too fast in the 1980s. And unlike the teen sex comedies that overshadowed it (Porky's, Fast Times at Ridgement High), the movie holds up well against time...\\
\textbf{Output:}\\
Positive\\
\textbf{Training Format:} \\
Training input = \textbf{Prompt} + \textbf{Input},  Training label = \textbf{Output} \\
\hline
\textbf{Dialog} \\
\hline
\textbf{Instruction:} \\If you are a doctor, please answer the medical questions based on the patient's description.\\
\textbf{Input:} \\I woke up this morning feeling the whole room is spinning when i was sitting down. I went to the bathroom walking unsteadily, as i tried to focus i feel nauseous. I try to vomit but it wont come out.. After taking panadol and sleep for few hours, i still feel the same.. \\
\textbf{Output:} \\Hi, Thank you for posting your query. The most likely cause for your symptoms is benign paroxysmal positional vertigo (BPPV), a type of peripheral vertigo. In this condition, the most common symptom is dizziness or giddiness, which is made worse with movements. ...\\
\textbf{Training Format:} \\
Training input = \textbf{Instruction} + \textbf{Input},  Training label = \textbf{Output}\\
\hline
\textbf{Question and answering} \\
\hline
\textbf{Question:}\\
Who was the Norse leader?\\
\textbf{Input:} \\
... They were descended from Norse ("Norman" comes from "Norseman") raiders and pirates from Denmark, Iceland and Norway who, under their leader Rollo, agreed to swear fealty to King Charles III of West Francia. ...\\
\textbf{Output:}\\
Rollo\\
\textbf{Training Format:} \\
Training input = \textbf{Question} + \textbf{Input},  Training label = \textbf{Output} \\
\hline
\end{tabular}

\label{tab: summ}
\end{table*}

\section{Disparate Memorization on GPT-Neo}
\label{gptneo-ft}

We also fine-tuned decoder-only GPT-Neo-125m models and also observed similar findings with T5-base, which suggests our findings are generalizable. The results are reported in Table \ref{tab:self fine-tune gptneo}. We can clearly observe that the memorization increase for summarization and dialog is much more significant than QA and sentiment classification. 

\begin{table*}[htbp]
\centering
\caption{Memorization rate of gpt-neo-125m fine-tuned on various tasks}
\label{tab:self fine-tune gptneo}
\resizebox{0.45\textwidth}{!}{
\begin{tabular}{@{}c|c|c|c@{}}
\toprule
Task & Dataset & Model & \begin{tabular}[c]{@{}c@{}} Total \\Mem Rate \end{tabular} \\
\midrule
\multirow{3}{*}{Summarization} & \multirow{3}{*}{Multi-news} & GPT-Neo & 25.2\% \\
 & & GPT-Neo-ft & 44.3\% \\
 & & Difference & $\uparrow$19.1\% \\
\midrule
\multirow{3}{*}{Dialog} & \multirow{3}{*}{chatdoctor} & GPT-Neo & 2.5\% \\
 & & GPT-Neo-ft & 7.8\% \\
 & & Difference & $\uparrow$5.3\% \\
\midrule
\multirow{3}{*}{\begin{tabular}[c]{@{}c@{}} Sentiment \\ Classification \end{tabular}} & \multirow{3}{*}{imdb} & GPT-Neo & 3.9\% \\
 & & GPT-Neo-ft & 4.2\% \\
 & & Difference & $\uparrow$0.3\% \\
\midrule
\multirow{3}{*}{\begin{tabular}[c]{@{}c@{}} Reading \\ Comprehension \end{tabular}} & \multirow{3}{*}{Squad\_v2} & GPT-Neo & 0.02\% \\
 & & GPT-Neo-ft & 0.04\% \\
 & & Difference & $\uparrow$0.02\% \\
\midrule
\multirow{3}{*}{Translation} & \multirow{3}{*}{wmt} & GPT-Neo & 0.00\% \\
 & & GPT-Neo-ft & 0.00\% \\
 & & Difference & - \\
\bottomrule
\end{tabular}}
\end{table*}

\section{Ablation Studies}

\subsection{Training Epochs}
\label{ablation_training}
 Here, we present the memorization rates observed in checkpoints of our fine-tuned models across different epochs, as shown in Table~\ref{tab: ablation_epochs}.
 
 In our practice, we find that for low-memorization tasks like sentiment classification, no matter whether the model is well-trained, the memorization ratio remains low. However, for high-memorization tasks like dialog, if the model is not well-trained, the memorization will be low. So in our experiment, to make sure that our finetuned model is well-trained, we let the finetuned model have comparable performance with Flan-T5 as Flan-T5 is a well-trained model and has good performance on various tasks.

\subsection{Sampling Methods}
\label{decoding_ab}

We conduct ablation studies on different decoding methods in Table \ref{tab: t5_decoding}. From the results, we can find that:
\begin{itemize}
    \item For high-memory tasks such as summarization and Dialog,  sampling can reduce the memorization Rate and change the category distribution of memory samples.
   \item For low-memory tasks such as emotion classification, sampling does not significantly affect the memorization results.
   \item Irrespective of the decoding methodology employed, \textbf{a pronounced disparity in memorization across different tasks persists.} This suggests an inherent task-specific propensity towards memorization that is not substantially mitigated by variations in sampling techniques.
\end{itemize}

\subsection{Prefix Lengths}
\label{prefix_ab}

Here we change different prefix lengths of inputs and report the results in table \ref{tab: t5_prefix_length}. We include 2 high-memorization tasks(summarization and dialog) and 1 low-memorization task (sentiment classification). From the results we can observe that:
\begin{itemize}
    \item \textbf{The length of prefix tokens can affect memorization.} The length of prefix tokens does indeed impact memorization. Specifically, for summarization and Dialog tasks, the memorization Rate generally increases with the length of the prefix. This finding aligns with previous research on pre-trained memorization. However, for sentiment classification, changing the prefix does not result in significant changes, and increasing the prefix length does not necessarily lead to an increase in the memorization Rate.
    \item \textbf{The task disparity still exists when using different prefixes. }Furthermore, it is worth noting that despite the influence of different prefixes on memorization, there still exists a noticeable disparity in memorization across tasks. Therefore, our conclusion remains even using different prefixes.
\end{itemize}

\section{Statistical Significance Testing}
\label{Statistic Test}
In this section, the results displayed in Tables \ref{tab:Open_Sourced_test} to \ref{tab: t5_prefix_test} are derived from the data in Tables \ref{tab:Open_Sourced finetune}, \ref{tab:self finetune}, \ref{tab: t5_decoding}, \ref{tab: t5_prefix_length}, and Figure \ref{fig:Ablation}. Specifically, we conducted 1000 bootstrap experiments based on these sources to calculate confidence intervals at the 5\% and 95\% levels for the results presented in the aforementioned tables.
\begin{table*}[t]
\centering
\caption{Statistical significance test of open-sourced LLMs fine-tuned on various tasks}
\label{tab:Open_Sourced_test}
\resizebox{0.6\textwidth}{!}{
\begin{tabular}{@{}c|ccc@{}}
\toprule
Task & Dataset & Source Model & Total Mem Rate (CI) \\
\midrule
\href{https://huggingface.co/facebook/bart-large-cnn}{Summarization} & CNN/Daily Mail & Bart\_Large & [20.30\%, 21.10\%] \\
\href{https://huggingface.co/Narrativaai/BioGPT-Large-finetuned-chatdoctor}{Medical Dialog} & ChatDoctor & BioGPT & [18.86\%, 20.36\%] \\
\href{https://huggingface.co/potsawee/t5-large-generation-squad-QuestionAnswer}{Extractive QA} & SQuAD\_v2 & T5\_large & [0.05\%, 0.17\%] \\
\href{https://huggingface.co/potsawee/t5-large-generation-race-QuestionAnswer}{Abstractive QA} & Race & T5\_large & [0.19\%, 0.41\%] \\
\href{https://huggingface.co/facebook/wmt19-en-de}{Translation} & WMT\_19 & FSMT & [0\%, 0\%] \\
\href{https://huggingface.co/mrm8488/t5-base-finetuned-imdb-sentiment}{Sentiment Classification} & IMDB & T5-base & [0\%, 0\%] \\
\bottomrule
\end{tabular}
}
\end{table*}

\begin{table*}[htbp]
\centering
\caption{Statistical significance test of T5-base fine-tuned on various tasks}
\label{tab:self_finetune_test}
\resizebox{0.5\textwidth}{!}{
\begin{tabular}{@{}c|c|c|c@{}}
\toprule
Task & Dataset & Model & Total Mem Rate (CI) \\
\midrule
\multirow{3}{*}{Summarization} & \multirow{3}{*}{Multi-news} & T5-base & [13.28\%, 14.68\%] \\
 & & T5-finetuned & [21.52\%, 23.14\%] \\
 & & Difference & [7.84\%, 8.86\%] \\
\midrule
\multirow{3}{*}{Dialog} & \multirow{3}{*}{chatdoctor} & T5-base & [1.36\%, 1.84\%] \\
 & & T5-finetuned & [7.75\%, 8.81\%] \\
 & & Difference & [6.15\%, 7.19\%] \\
\midrule
\multirow{3}{*}{\begin{tabular}[c]{@{}c@{}}Sentiment\\Classification\end{tabular}} & \multirow{3}{*}{imdb} & T5-base & [0.61\%, 0.95\%] \\
 & & T5-finetuned & [0.62\%, 0.98\%] \\
 & & Difference & [0\%, 0.04\%] \\
\midrule
\multirow{3}{*}{\begin{tabular}[c]{@{}c@{}}Reading\\Comprehension\end{tabular}} & \multirow{3}{*}{Squad\_v2} & T5-base & [0.07\%, 0.09\%] \\
 & & T5-finetuned & [0.07\%, 0.23\%] \\
 & & Difference & [0.01\%, 0.19\%] \\
\midrule
\multirow{3}{*}{\begin{tabular}[c]{@{}c@{}}Translation\end{tabular}} & \multirow{3}{*}{wmt} & T5-base & [0\%, 0\%] \\
 & & T5-finetuned & [0\%, 0\%] \\
 & & Difference & [0\%, 0\%] \\
\bottomrule
\end{tabular}}
\end{table*}

\begin{table*}[htbp]
\centering
\caption{Statistical significant test of fine-tuned T5 with various sampling methods.}
\label{tab: t5_decoding_test}
\resizebox{0.6\textwidth}{!}{%
\begin{tabular}{c|c|c|c}
\toprule
Task & Dataset & Decoding & Total Mem Rate (CI) \\ 
\midrule
\multirow{4}{*}{Summarization} & \multirow{4}{*}{Multi\_news} & Top-K & [10.17\%, 11.43\%] \\
 & & Top-p & [12.89\%, 14.25\%] \\
 & & Temp & [5.37\%, 6.27\%] \\
 & & Greedy & [21.48\%, 23.18\%] \\
\midrule
\multirow{4}{*}{Dialog} & \multirow{4}{*}{HealthCareMagic} & Top-K & [5.32\%, 6.23\%] \\
 & & Top-p & [6.78\%, 7.77\%] \\
 & & Temp & [3.34\%, 4.08\%] \\
 & & Greedy & [7.75\%, 8.79\%] \\
\midrule
\multirow{4}{*}{Sentiment} & \multirow{4}{*}{IMDB} & Top-K & [0.83\%, 1.23\%] \\
 & & Top-p & [0.87\%, 1.29\%] \\
 & & Temp & [0.71\%, 1.08\%] \\
 & & Greedy & [0.63\%, 1.00\%] \\
\midrule
\multirow{4}{*}{QA} & \multirow{4}{*}{Squad\_v2} & Top-K & [0.08\%, 0.23\%] \\
 & & Top-p & [0.07\%, 0.23\%] \\
 & & Temp & [0.08\%, 0.23\%] \\
 & & Greedy & [0.08\%, 0.23\%] \\
\bottomrule
\end{tabular}
}
\end{table*}

\begin{table*}[htbp]
\centering
\caption{Statistical significance test of fine-tuned T5 with varying prefix lengths.}
\label{tab: t5_prefix_test}
\resizebox{0.6\textwidth}{!}{%
\begin{tabular}{c|c|c|c}
\toprule
Task & Dataset & Prefix length & Total Mem Rate (CI) \\
\midrule
\multirow{4}{*}{Summarization} & \multirow{4}{*}{Multi\_news} & 10 & [11.94\%, 12.56\%] \\
 &  & 30 & [20.27\%, 21.09\%] \\
 &  & 50 & [21.99\%, 22.67\%] \\
 &  & 100 & [29.22\%, 30.10\%] \\
\midrule
\multirow{3}{*}{Dialog} & \multirow{3}{*}{HealthCareMagic} & 10 & [5.79\%, 6.77\%] \\
 &  & 30 & [7.23\%, 8.29\%] \\
 &  & 50 & [7.76\%, 8.78\%] \\
\midrule
\multirow{4}{*}{Sentiment} & \multirow{4}{*}{IMDB} & 10 & [1.13\%, 1.61\%] \\
 &  & 30 & [0.96\%, 1.40\%] \\
 &  & 50 & [0.63\%, 0.97\%] \\
 &  & 100 & [1.15\%, 1.63\%] \\
\midrule
\multirow{4}{*}{QA} & \multirow{4}{*}{Squad\_v2} & 10 & [0.09\%, 0.27\%] \\
 &  & 30 & [0.06\%, 0.22\%] \\
 &  & 50 & [0.07\%, 0.24\%] \\
 &  & 100 & [0.06\%, 0.21\%] \\

\bottomrule
\end{tabular}
}
\end{table*}

\color{black}
\section{Dataset and Model used}

\label{Datasets and model}
\paragraph{Datasets} 
In our study, we utilized various datasets for preliminary experiments and model fine-tuning. For the summarization task, we used CNN/Daily Mail in the preliminary phase, a dataset comprising 287k training rows and 10k evaluation rows. For fine-tuning, we employed the \textbf{Multi-News} dataset \cite{alex2019multinews}, which includes news articles and summaries, using a 45k training set and a 5.62k test set. The \textbf{IMDB} dataset \cite{maas-EtAl:2011:ACL-HLT2011} was used for binary sentiment classification, consisting of 25k training and 25k test movie reviews. In the dialog task, we use \href{https://huggingface.co/datasets/lavita/ChatDoctor-HealthCareMagic-100k}{\textbf{HealthcareMagic}} dataset, comprising 112k training rows and a 12k test set. For extractive QA, we utilized the \textbf{SQuAD} v2 dataset \cite{2016arXiv160605250R}, featuring questions based on Wikipedia articles, with 130k training and 11.9k test rows. The translation task involved a preliminary study using \href{https://huggingface.co/datasets/wmt19}{\textbf{WMT19}} and fine-tuning on an English-to-German subset of \href{https://huggingface.co/datasets/wmt16}{\textbf{WMT16}}, with 450.87k training and 3k test rows. Finally, for the controlling experiment, we used the \textbf{RentTheRunway} dataset \cite{10.1145/3240323.3240398}, containing clothing review data, with 111k training and 12k test rows. This dataset was used for fine-tuning both the summarization model and the binary sentiment classification task.

\paragraph{Models}
In the preliminary study, we consider Bart-Large from Bart family, T5-base and T5-large from T5 family, FSMT (FairSeq MachineTranslation), and BioGPT. For our self-fine-tuned models, we select T5-base architecture from the T5 family for all experiments.

\paragraph{Fine-tuned Methods}
In the fine-tuning process, we consider all the tasks as generation tasks and use the format of instruction tuning. Here we provide fine-tuned templates of different tasks in Table \ref{training template}
\section{Performance of Self-fine-tuned Models}
\label{finetune performance}
We finetune the T5-base model to achieve better or comparable performance with the Google fine-tuned public model FLAN-T5. In Table~\ref{tab:summary_performance}, we show the performance of the summarization task. Our fine-tuned model achieves a similar rouge score with FLAN-T5. In Table~\ref{tab:Sentiment_performance}, We show that the accuracy of our model is better than FLAN-T5 regarding binary sentiment classification. For Dialog task, Our model performance much better than FLAN-T5 as shown in Table~\ref{tab:dialogue}. For the Extractive question-answering task, we fine-tune the model in a sequence-to-sequence learning form while we evaluate the exact match of the answer term. Results are shown in Table~\ref{tab:QA}. For the RentTheRunway fine-tuning experiment, we present the results in Table \ref{tab:renttherunway} and Table \ref{tab:renttherunway1}.
\begin{table*}[htbp!]
\centering
\caption{Summarization}
\label{tab:summary_performance}
\resizebox{0.65\textwidth}{!}
{
\begin{tabular}{c|c|c|c|c|c}
\toprule
Dataset          & Model   & Rouge1      & Rouge2                & RougeL       & RougeLSum                \\ 
\midrule
multi\_news & FLAN-T5-small&  0.264 & 0.092 & 0.168 & 0.168\\
multi\_news & Our fine-tuned T5-small& 0.308 & 0.088 & 0.187 & 0.187\\
multi\_news & FLAN-T5-base&0.291&0.098&0.237&0.237\\
multi\_news & Our fine-tuned T5-base&0.298&0.103&0.201&0.201\\
multi\_news & FLAN-T5-large& 0.256 & 0.087 & 0.165 & 0.165\\
multi\_news & Our fine-tuned T5-large& 0.368& 0.122 & 0.218 & 0.218\\
multi\_news & FLAN-T5-3b& 0.264 & 0.092 & 0.232 & 0.232\\
multi\_news & Our fine-tuned T5-3b & 0.387 & 0.136 & 0.168 & 0.168\\

\bottomrule
\end{tabular}}
\vspace{0.3cm}
\centering
\caption{Sentiment classification}
\label{tab:Sentiment_performance}
\resizebox{0.4\textwidth}{!}{
\begin{tabular}{c|c|c}
\toprule
Dataset          & Model   & Accuracy(\%)  \\ 
\midrule
IMDB & FLAN-T5-small & 94.17\\
IMDB & Our fine-tuned T5-small& 95.30 \\
IMDB & FLAN-T5-base & 93.56\\
IMDB & Our fine-tuned T5-base& 94.64 \\
IMDB & FLAN-T5-large & 94.50\\
IMDB & Our fine-tuned T5-large& 95.30\\
IMDB & FLAN-T5-3b & 97.10 \\
IMDB & Our fine-tuned T5-3b &  95.20\\
\bottomrule
\end{tabular}}%
\vspace{0.2cm}
\caption{Dialog}
\label{tab:dialogue}
\resizebox{0.7\textwidth}{!}{
\begin{tabular}{c|c|c|c|c|c}
\toprule
 Dataset     & Model   & Rouge1     & Rouge2      & RougeL       & RougeLSum   \\ \midrule
HealthCareMagic& FLAN-T5-small & 0.041 & 0.004 & 0.031 & 0.031\\
HealthCareMagic& Our fine-tuned T5-small & 0.131 & 0.063 & 0.154 & 0.154\\
HealthCareMagic& FLAN-T5-base & 0.055 &0.006 & 0.039 & 0.039 \\
HealthCareMagic& Our fine-tuned T5-base& 0.298&0.103&0.201&0.201\\
HealthCareMagic&FLAN-T5-large & 0.068 & 0.007 & 0.050 & 0.050\\
HealthCareMagic& Our fine-tuned T5-large & 0.220 & 0.063 & 0.154 & 0.154\\
HealthCareMagic&FLAN-T5-3b & 0.073 & 0.010 & 0.055 & 0.055\\
HealthCareMagic& Our fine-tuned T5-3b & 0.139 & 0.012 & 0.094 & 0.094\\
\bottomrule
\end{tabular}}%


\centering
\caption{Question answering}
\label{tab:QA}
\resizebox{0.45\textwidth}{!}{
\begin{tabular}{c|c|c}
\toprule
Dataset      & Model   & Exact Match(\%) \\ \midrule
SQuAD v2 & FLAN-T5-small & 35.23\\
SQuAD v2 & Our finetuned T5-small & 49.1 \\
SQuAD v2 & FLAN-T5-base & 34.30 \\
SQuAD v2 & Our finetuned T5-base & 44.00\\
SQuAD v2 & FLAN-T5-large & 43.36\\
SQuAD v2 & Our finetuned T5-large & 53.20\\
SQuAD v2 & FLAN-T5-3b & 44.60\\
SQuAD v2 & Our finetuned T5-3b & 37.20\\
\bottomrule
\end{tabular}}%
\end{table*}
\begin{table*}
\vspace{0.2cm}
\centering
\caption{Multi task trained with RentTheRunway}
\label{tab:renttherunway}
\resizebox{0.65\textwidth}{!}{
\begin{tabular}{c|c|c|c|c|c}
\toprule
 Dataset          & Model   & Rouge1      & Rouge                  & RougeL       & RougeSum        \\ 
 \midrule
Summary & FLAN-T5-Base & 0.1743 & 0.0436 & 0.1598 & 0.1598  \\
Summary & Our finetuned T5-Base & 0.1743 & 0.0436 & 0.1598 & 0.1598  \\
\bottomrule
\end{tabular}}
\vspace{0.2cm}
\centering
\caption{Multi task trained with RentTheRunway}
\label{tab:renttherunway1}
\resizebox{0.55\textwidth}{!}{
\begin{tabular}{c|c|c}
\toprule
 Dataset          & Model   & Accuracy(\%)     \\ 
 \midrule
Sentiment Classification & Flan T5-base & 86.60 \\
Sentiment Classification & Our fine-tuned T5-base & 98.07 \\
\bottomrule
\end{tabular}}
\end{table*}
\section{Memorized Examples}
\label{Mem Cases}
We present memorization examples of verbatim, paraphrase and idea plagiarism of different models in Table \ref{tab:real_examples2}.
\begin{table*}[!htbp]
\vspace{0cm}
\centering
\caption{Examples of memorization cases. Duplicated texts are highlighted with yellow marks. Personally identifiable information (PII) and other words that may lead to privacy concern in generated text are masked as red.
}
\setlength{\leftskip}{-50pt}

\setlength{\tabcolsep}{2.5pt}
\begin{tabular}{c|cc}
\toprule
Type & Machine-Written Text & Training Text \\
\midrule
\multirow{2}*{Paraphrase} 
&  \begin{tabular}[c]{@{}c@{}} \\At least\highlightyellow{10 people} and two attackers were \highlightyellow{killed} \\in Tuesday's attack against the luxurious \\ \highlightred{Corinthia Hotel in Tripoli, Libya}, a spokesman for \\ a security division of the Ministry of Interior in \\ \highlightyellow{Tripoli} said. \highlightyellow{Five foreigners -- one American} \\(\textit{\textbf{Summarization: CNN Daily, Bart Large}}) \\ \newline \end{tabular} &  \begin{tabular}[c]{@{}c@{}}On January 27, gunmen claiming to be affiliated \\with ISIS attacked \highlightyellow{the Corinthia Hotel in Tripoli,}\\ \highlightyellow{Libya, } which is favored by government officials and \\foreigners. They \highlightyellow{killed 10 people} after storming into \\ the lobby and firing guns at hotel guests. Five \\of the victims were \highlightyellow{foreigners, one an American}.  \end{tabular}  \\  
\midrule
\multirow{2}*{Paraphrase} &  \begin{tabular}[c]{@{}c@{}} \\-- \highlightred{Argentine President Cristina Fernandez de Kirchner}\\ was told \highlightyellow{to take a month off work} after doctors \\diagnosed her with a subdural \highlightyellow{hematoma}. \\ (\textit{\textbf{Summarization: CNN Daily, Bart Large}}) \\ \newline \end{tabular} 
 &  \begin{tabular}[c]{@{}c@{}}...\highlightyellow{hematoma and} \\ \highlightyellow{said she needed to take a month off of work.}\end{tabular} \\  
\midrule
\multirow{2}*{Paraphrase} 
 &  \begin{tabular}[c]{@{}c@{}}\\ I am a 20-year-old guy \highlightyellow{20 years old. I have been}\\ \highlightred{(***harassment word***)} for a long time. \\(\textit{\textbf{Dialog: 
 HealthCareMagic, Finetuned T5}}) \\ \newline \end{tabular}&  \begin{tabular}[c]{@{}c@{}}...\highlightyellow{20 years old  i have been}. \\ \highlightyellow{(***harassment word***) }regularly \highlightyellow{for} past 5 years \end{tabular} \\
 \midrule
 
\multirow{2}*{Paraphrase} &  \begin{tabular}[c]{@{}c@{}} \\ \highlightyellow{The Chargers responded with} a 1-\highlightyellow{yard TD}\\ run by \highlightyellow{RB} \highlightred{LaDainian Tomlinson}. \\(\textit{\textbf{Abstractive QA: Finetuned T5}}) \\ \newline \end{tabular}
 &  \begin{tabular}[c]{@{}c@{}} \highlightyellow{The Chargers} would respond \highlightyellow{with RB} \\ \highlightyellow{LaDainian Tomlinson with a} 4-\highlightyellow{yard TD run.}\\ \end{tabular} \\
\midrule

\multirow{2}*{Paraphrase} 
&  \begin{tabular}[c]{@{}c@{}} \\ – President Trump has \highlightyellow{named} \highlightred{Mick Mulvaney}\\ to replace \highlightred{John Kelly}, \highlightyellow{the White House chief} \\of staff who left the White House in December." \\ (\textbf{Summary: Multi news, Finetuned T5}) \\ \newline \end{tabular} & \begin{tabular}[c]{@{}c@{}} \\ “I am pleased to announce that \highlightyellow{Mick Mulvaney}, \\Director of the Office of Management \& Budget,\\ will be \highlightyellow{named} Acting \highlightyellow{White House Chief of Staff}, \\replacing General \highlightyellow{John Kelly}, who has served our ...\end{tabular}   \\

\midrule

\multirow{2}*{Verbatim} &  \begin{tabular}[c]{@{}c@{}}\\ \highlightred{Rachel's son Liam} \highlightyellow{in a house}\\ \highlightyellow{near} \highlightred{Glenrothes} \highlightyellow{on 22 March 2014. }\\(\textit{\textbf{Summary: CNN Daily, Finetuned T5}}) \\ \newline \end{tabular} 
 & \begin{tabular}[c]{@{}c@{}}\highlightyellow{Rachel's son Liam in a house}\\ \highlightyellow{near Glenrothes on 22 March 2014.}  \end{tabular} \\
 \midrule
  
 \multirow{2}*{Verbatim} &  \begin{tabular}[c]{@{}c@{}}\\ \highlightyellow{divided Wednesday during heated arguments}\\ \highlightyellow{over President} \highlightred{Obama's} \highlightyellow{health care law, but}\\(\textit{\textbf{Summary: Multi news, Finetuned T5}}) \\ \newline\end{tabular} 
 & \begin{tabular}[c]{@{}c@{}} \highlightyellow{divided Wednesday during heated arguments} \\ \highlightyellow{over President Obama's health care law, but}\end{tabular} \\
 \midrule
\multirow{2}*{Verbatim} &  \begin{tabular}[c]{@{}c@{}}\\ \highlightyellow{and liver cirrhosis in dec 2011} \\ \highlightyellow{modified akt staarted because of cirrhosis i.e} \\ (\textit{\textbf{Dialog:Finetuned T5, ChatDoctor}})\\ \end{tabular} 
 & \begin{tabular}[c]{@{}c@{}}\highlightyellow{and liver cirrhosis in dec 2011} \\ \highlightyellow{modified akt staarted because of cirrhosis i.e  }\end{tabular} \\
\bottomrule
\end{tabular}
\label{tab:real_examples2}

\end{table*}

\begin{table*}[!htbp]
\centering
\setlength{\leftskip}{-55pt}

\setlength{\tabcolsep}{2pt}
\begin{tabular}{c|cc}

\toprule
Type & Machine-Written Text & Training Text \\
\midrule

\multirow{2}*{Verbatim} 
 &  \begin{tabular}[c]{@{}c@{}}\\ \highlightred{River Martinez}, \highlightyellow{10, breaks camp at the Upper} \\ \highlightyellow{Pines Campground in Yosemite National Park, Calif.,}\\ \highlightyellow{on Wednesday, July 25, 2018.}\\ (\textit{\textbf{Summary: Multi news, Finetuned T5
}})\\\newline \end{tabular}&  \begin{tabular}[c]{@{}c@{}}  \highlightyellow{River Martinez, 10, breaks camp at the Upper}\\ \highlightyellow{Pines Campground in Yosemite National Park, Calif.,}\\ \highlightyellow{on Wednesday, July 25, 2018.}\\ \end{tabular} \\
 \midrule
 
  \multirow{1}*{Verbatim} & \begin{tabular}[c]{@{}c@{}}\\ \highlightyellow{A rare blue lobster caught by local lobsterman, }\\ \highlightyellow{Greg Ward, is on display at the} \highlightred{Seacoast Science}\\\highlightred{ Center in Rye, N.H., on Tuesday, July 18, 2017.}\\(\textit{\textbf{Summary: Xsum, Finetuned T5}}) \\ \newline\end{tabular} 
 & \begin{tabular}[c]{@{}c@{}}\\ \highlightyellow{A rare blue lobster caught by local lobsterman, }\\  \highlightyellow{Greg Ward, is on display at the Seacoast Science }\\  \highlightyellow{Center in Rye, N.H., on Tuesday, July 18, 2017. }\\ \newline\end{tabular}  \\
 \midrule
  
 \multirow{2}*{Verbatim} &  \begin{tabular}[c]{@{}c@{}}\\ \highlightred{Sheffield} \highlightyellow{homered twice and keyed} \\ \highlightyellow{a four-run rally in the ninth} \\ \highlightyellow{inning Thursday night, sending the }\\(\textit{\textbf{Classification: AG news, Finetuned T5}}) \\ \newline \end{tabular}
 & \begin{tabular}[c]{@{}c@{}} \highlightyellow{Sheffield homered twice and keyed}\\ \highlightyellow{a four-run rally in the ninth}\\ \highlightyellow{inning Thursday night, sending the} \end{tabular} \\
 \midrule
  \multirow{2}*{Idea} & \begin{tabular}[c]{@{}c@{}}\\ \highlightred{KUALA LUMPUR (Reuters) - Kim Jong Un's} \\half-brother\highlightyellow{was carrying \$100,000 in cash in}\\ \highlightyellow{his backpack} at the time of his murder, the officer\\ investigating the case told a police officer" \\(\textit{\textbf{Summary: Multi news, Finetuned T5}}) \\ \newline\end{tabular} 
 & \begin{tabular}[c]{@{}c@{}}\\ Wan Azirul testified that \highlightyellow{Kim was carrying}\\ \highlightyellow{\$100,000 in cash in his backpack.}\\ \newline\end{tabular}  \\
  \midrule
 \multirow{3}*{Idea} &  \begin{tabular}[c]{@{}c@{}} \\ \highlightred{Alan Dawson, 64, of Urmston,} was convicted of \\seven counts of indecent assault and \highlightyellow{one count} \\ \highlightyellow{of rape at} Manchester Crown Court. \\(\textit{\textbf{Summary: Xsum, Finetuned T5}}) \\ \newline \end{tabular} 
 &  \begin{tabular}[c]{@{}c@{}} ...is charged with \highlightyellow{one count of rape} \\and \highlightyellow{one count} of sexual \highlightyellow{assault.} \end{tabular} \\
 \midrule
 \multirow{3}*{Idea} &  \begin{tabular}[c]{@{}c@{}} \\ – \highlightyellow{Trey Radel, the Florida Rep. who was arrested}\\ \highlightyellow{last month for buying cocaine}, is a freshman congressman \\who has been a big news story for the Washington Post.  \\(\textit{\textbf{Summary: Multi News, Finetuned T5}}) \\ \newline \end{tabular} 
 &  \begin{tabular}[c]{@{}c@{}} Post) \textbackslash n \textbackslash \highlightyellow{n Florida Rep. Trey Radel (R-Fla.)} \\ \highlightyellow{was arrested last month for buying cocaine.}\\\end{tabular} \\
 \midrule
\multirow{2}*{Idea} &  \begin{tabular}[c]{@{}c@{}} \\ \highlightred{Abdul Aziz} believes he was \highlightyellow{standing right next to a shooter} \\when gunmen opened fire at a parade in new orleans, \\injuring 19 people. "Everyone around me was \highlightyellow{right next} to a \\shooter," \highlightred{Abdul Aziz} said.\\ (\textit{\textbf{Summary: CNN Daily, Finetuned T5}}) \\ \newline \end{tabular} 
 &  \begin{tabular}[c]{@{}c@{}}  I \highlightyellow{was standing}, I \highlightyellow{believe}, \highlightyellow{right next to the shooter}.  \end{tabular} \\
\bottomrule
\end{tabular}
\label{tab:real_examples3}

\end{table*}

\section{Attention Maps}
\label{attention_ab}

\subsection{Attention Maps of Translation Tasks}
\label{attention_translation}
In Figure \ref{fig:heatmap-translation}, we present the attention maps for translation tasks. The visualized examples reveal that attention scores tend to focus on a select few input features corresponding to each output token. This concentration of attention is likely due to the nature of translation tasks, where the model typically does not require full detail attention but rather focuses on key tokens for accurate translation. As a result, attention is directed predominantly towards specific features, leading to reduced overall memorization. These observations are consistent with our findings from other tasks.

\begin{figure}[t]
\centering

\resizebox{\textwidth}{!}{%
    \begin{minipage}{\textwidth}
        \subfloat[Sample1-trans.]{\includegraphics[width=.25\textwidth]{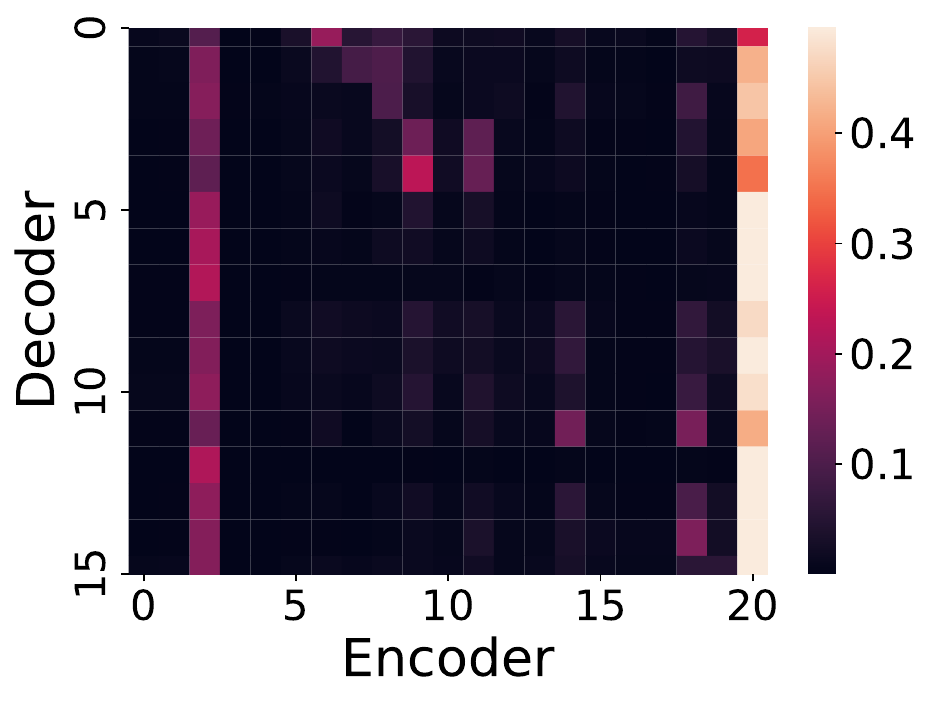}}
        \label{fig:translation-heatmap1}
        \subfloat[Sample2-trans.]{\includegraphics[width=.25\textwidth]{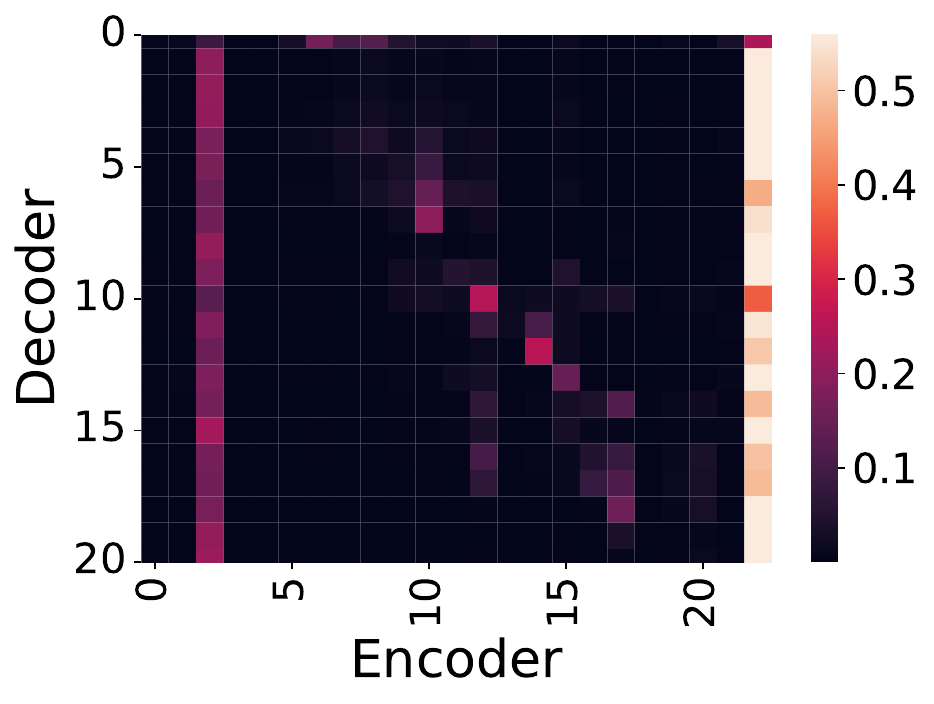}}
        \label{fig:translation-heatmap2}\\
        \subfloat[Sample3-trans.]{\includegraphics[width=.25\textwidth]{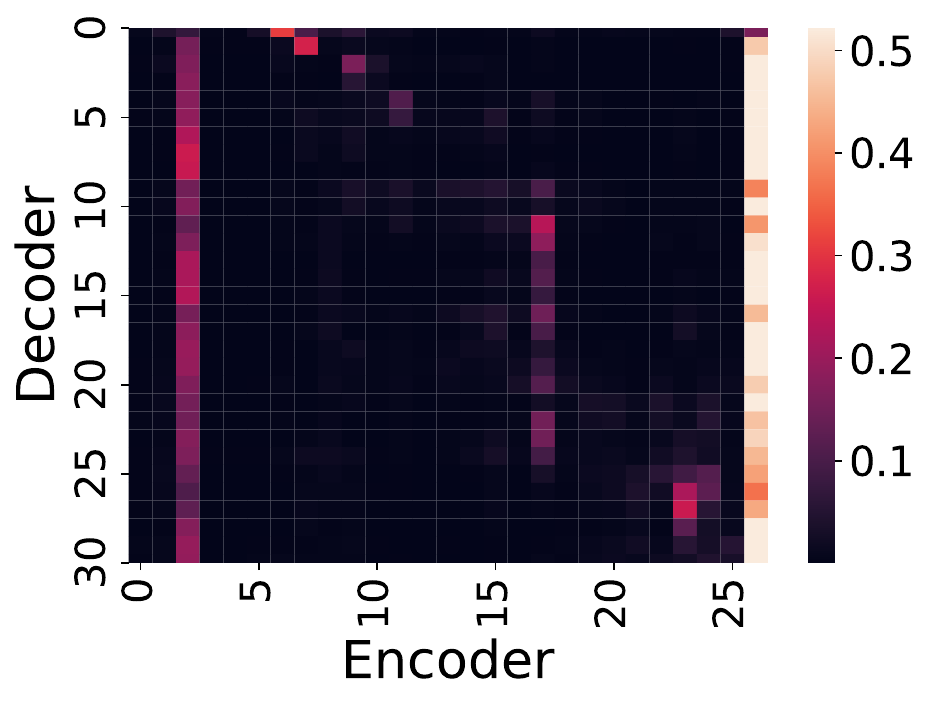}}
        \label{fig:translation-heatmap3}
        \subfloat[Sample4-trans.]{\includegraphics[width=.25\textwidth]{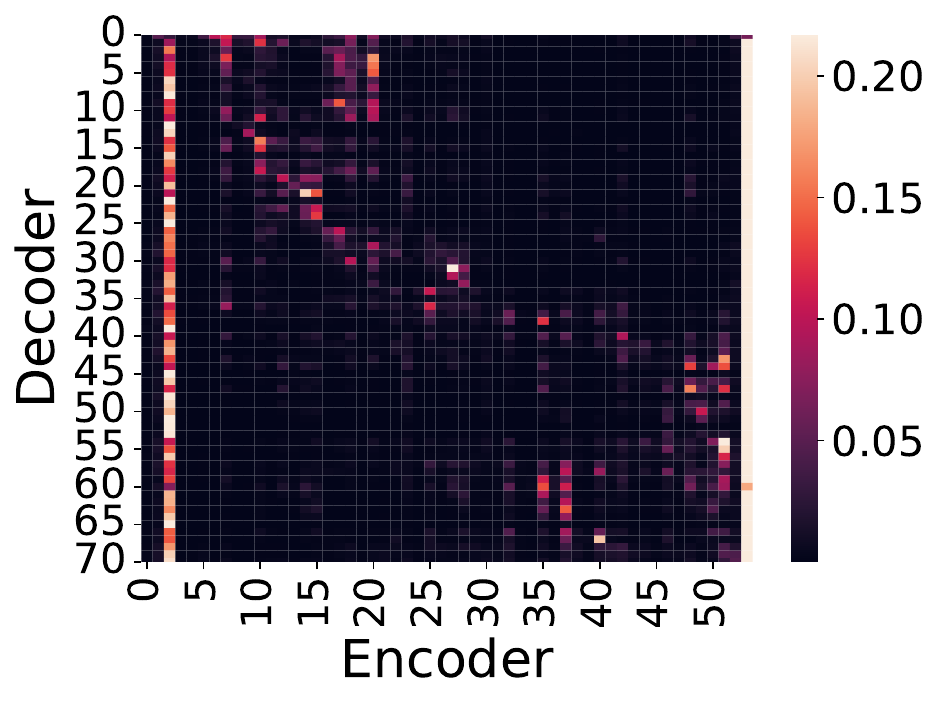}}
        \label{fig:translation-heatmap4}
        
    \end{minipage}
}

\caption{Decoder-encoder attention heatmaps on translation.}
\label{fig:heatmap-translation}
\end{figure}

\subsection{Attention Maps of T5-base when doing different tasks}
\label{attention_t5base}

To validate that attention patterns are more intrinsic properties of the tasks themself, we visualize the attention maps of the T5-base model(without fine-tuning) when doing different tasks in Figure \ref{fig:heatmap-base}. Specifically, we use the same instruction and input-output pairs of fine-tuning data as Section 5.1, but just change the model from finetuned-T5 to T5-base. From the Figure we can see that the disparity still exists across different tasks. And for each task, the attention patterns are similar to that of Fine-tuned T5. It further validates that the information needed to complete certain tasks is the intrinsic property of the task. 
\subsection{Attention Maps of different samples}
\label{sample attention}
In this section, we extend our visualization of attention maps across a broader range of samples and tasks, from Figure \ref{fig:heatmap-sample-summarization} to Figure \ref{fig:heatmap-sample-QA}. It is evident that memorization patterns differ significantly among tasks. Tasks with higher memorization requirements, such as summarization, display densely distributed attention scores, while those with lower memorization needs, like Extractive QA, exhibit more focused attention distributions.

\subsection{Attention Maps of Different encoder-decoder layers}
\label{attention_layers}
Here we visualize the attention maps of different encoder-decoder layers in Figure \ref{fig:heatmap-layer-summarization} to Figure \ref{fig:heatmap-layer-QA}. We can clearly observed consistent patterns across various layers of the encoder-decoder attention mechanism, with high memorization tasks showing dense attention and low memorization tasks focusing attention on fewer positions.
\clearpage
\begin{figure*}[t]
\centering
\resizebox{\textwidth}{!}{%
    \begin{minipage}{\textwidth}
        \subfloat[]{\includegraphics[width=.25\textwidth]{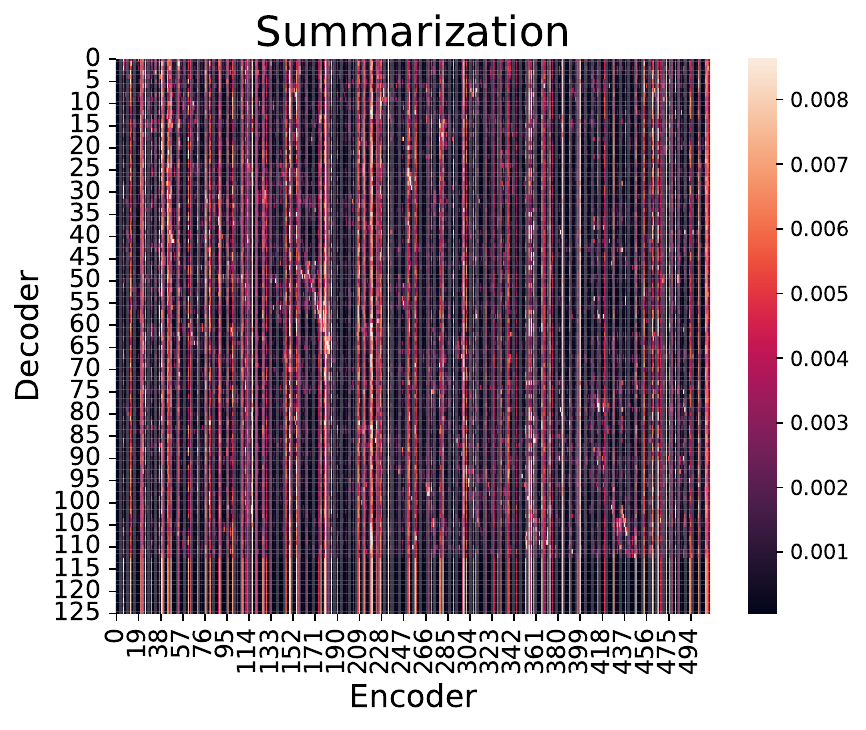}
        \label{fig:summary-base}}
        \subfloat[]{\includegraphics[width=.25\textwidth]{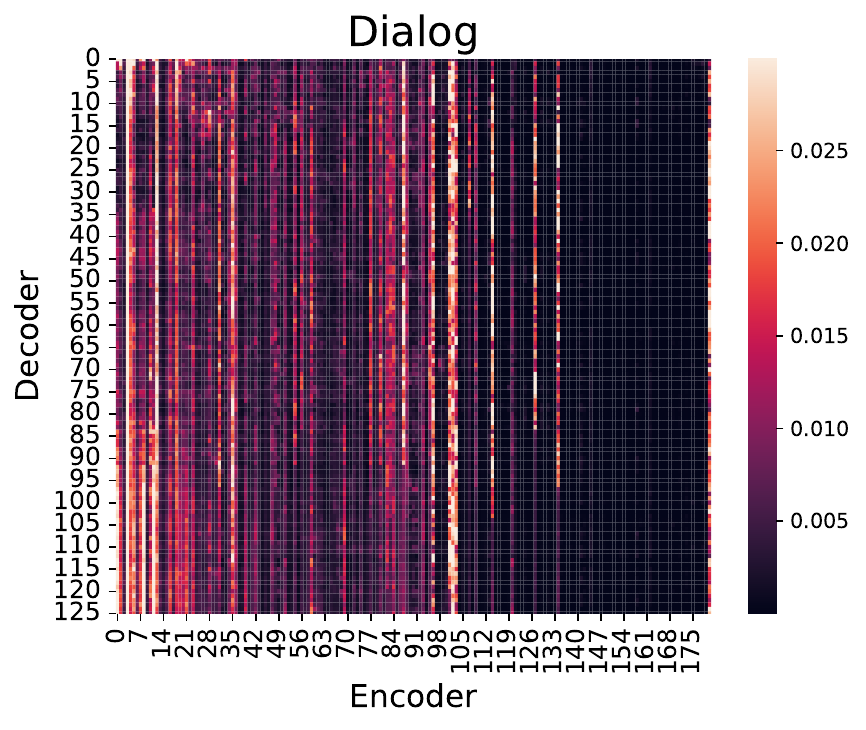}
        \label{fig:dialog-base}}
        \subfloat[]{\includegraphics[width=.25\textwidth]{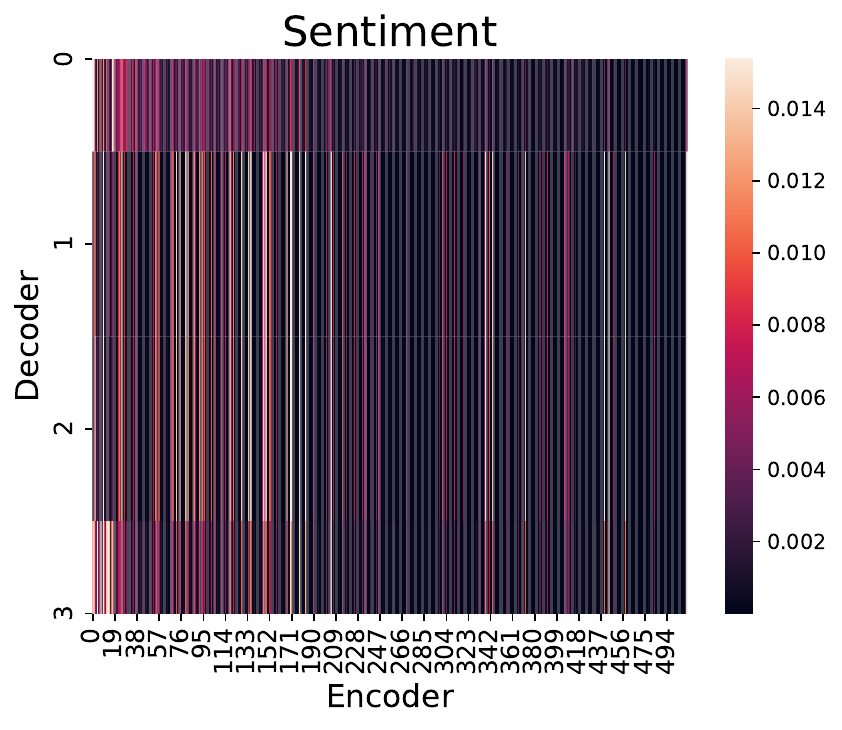}
        \label{fig:sentiment-base}}
        \subfloat[]{\includegraphics[width=.25\textwidth]{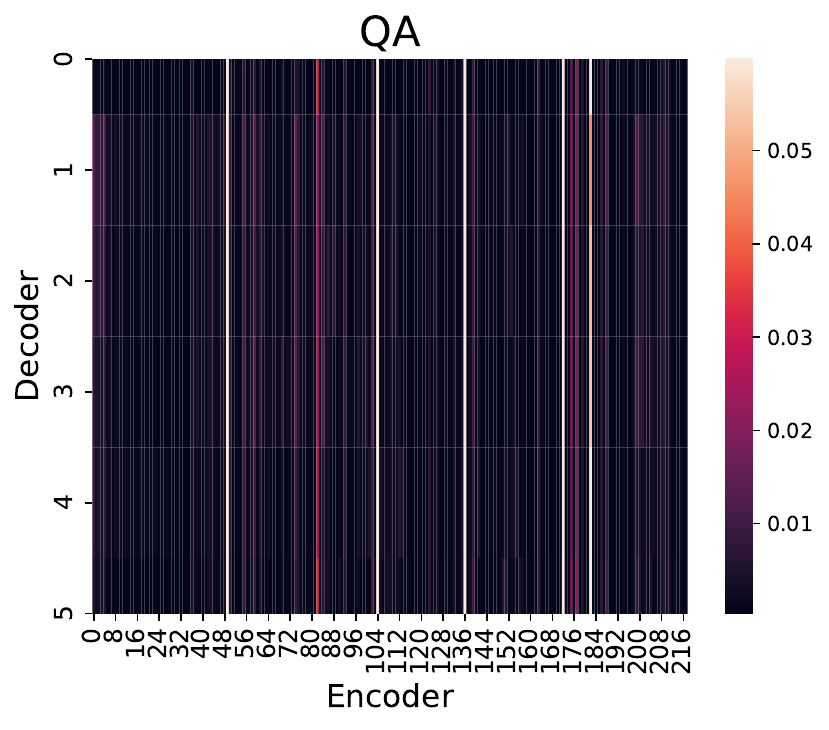}
        \label{fig:QA-base}}
    \end{minipage}
}

\caption{Average decode-encoder attention heatmaps on (a) summarization, (b) dialog, (c) sentiment, and (d) QA on T5-base across 10 samples}
\label{fig:heatmap-base}
\end{figure*}

\begin{figure*}[t]
\centering
\resizebox{\textwidth}{!}{%
    \includegraphics[]{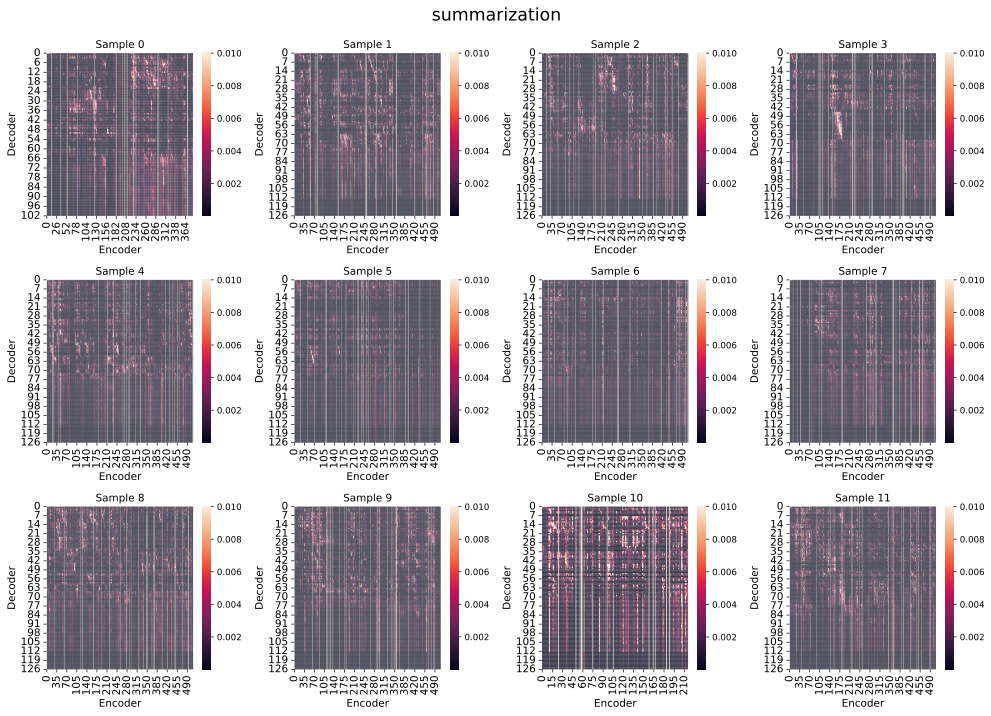}
    
}
\caption{Attention heatmaps of different samples on  summarization}
\label{fig:heatmap-sample-summarization}
\end{figure*}
\begin{figure*}[t]
\centering
\resizebox{\textwidth}{!}{%
    \includegraphics[]{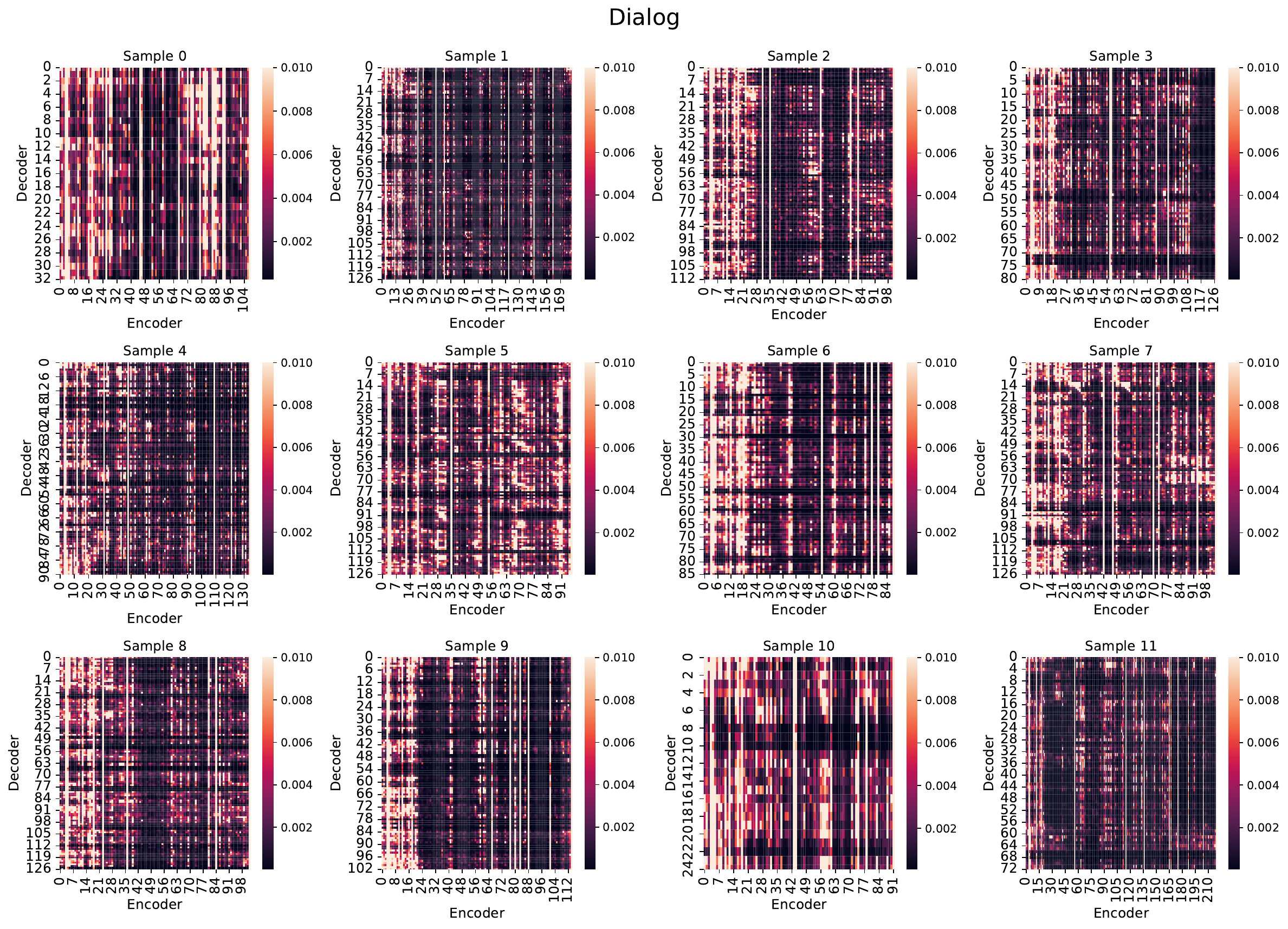}
    
}
\caption{Attention heatmaps of different samples on  dialog}
\label{fig:heatmap-sample-dialog}
\end{figure*}
\begin{figure*}[t]
\centering
\resizebox{\textwidth}{!}{%
    \includegraphics[]{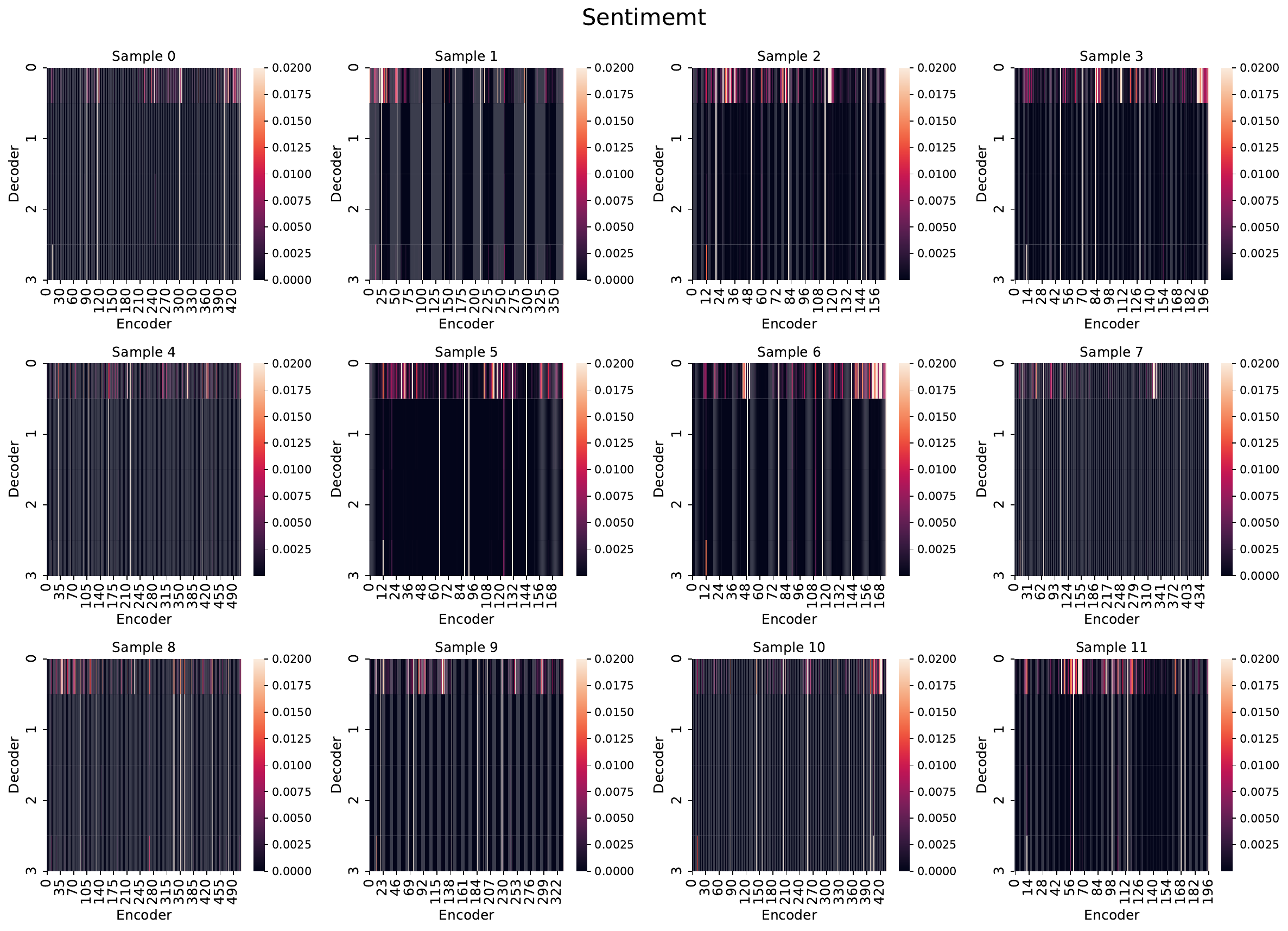}
    
}
\caption{Attention heatmaps of different samples on  sentiment classification}
\label{fig:heatmap-sample-sentiment}
\end{figure*}
\begin{figure*}[t]
\centering
\resizebox{\textwidth}{!}{%
    \includegraphics[]{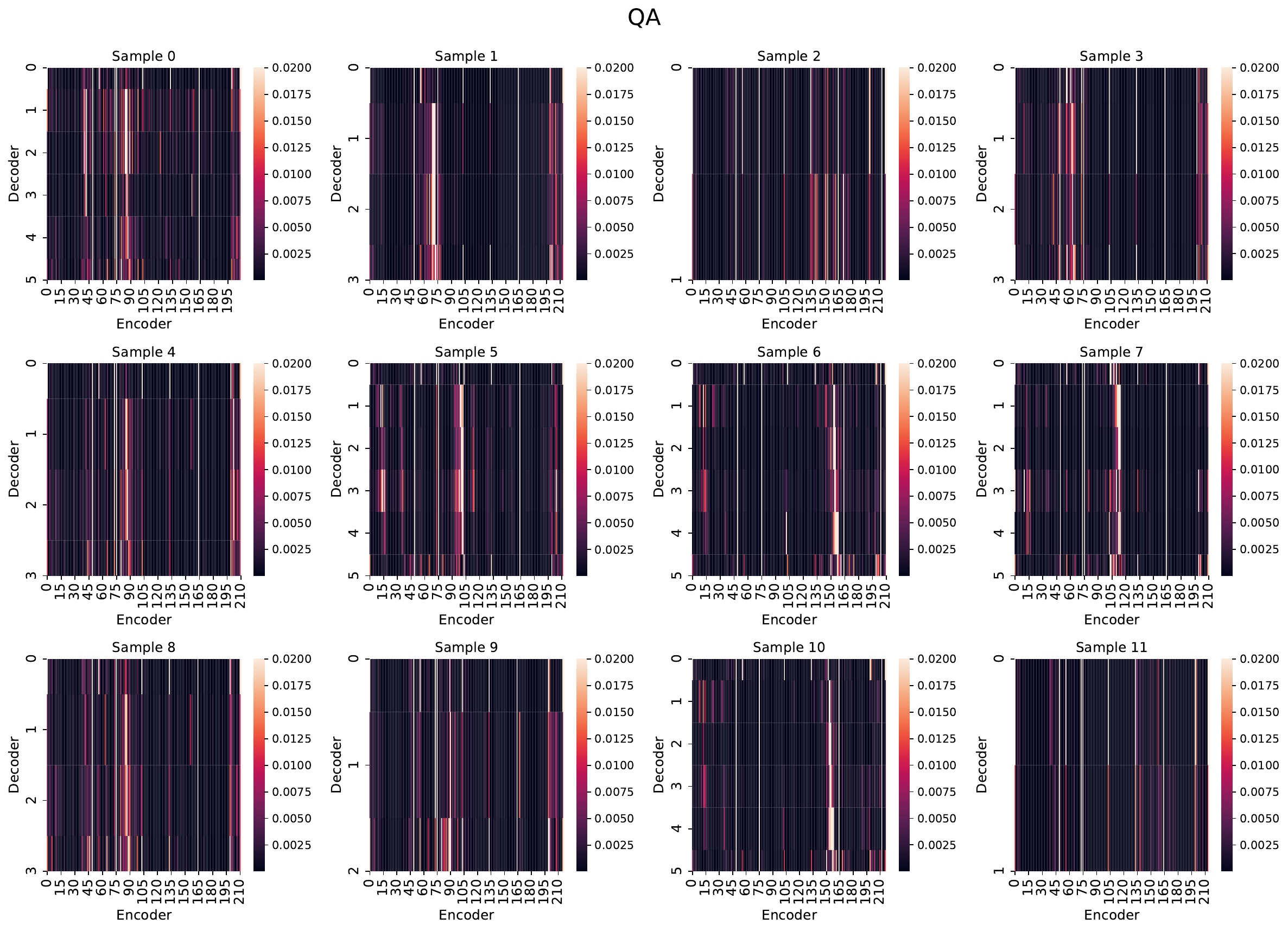}
    
}
\caption{Attention heatmaps of different samples on  QA}
\label{fig:heatmap-sample-QA}
\end{figure*}
\begin{figure*}[t]
\centering
\resizebox{\textwidth}{!}{%
    \includegraphics[]{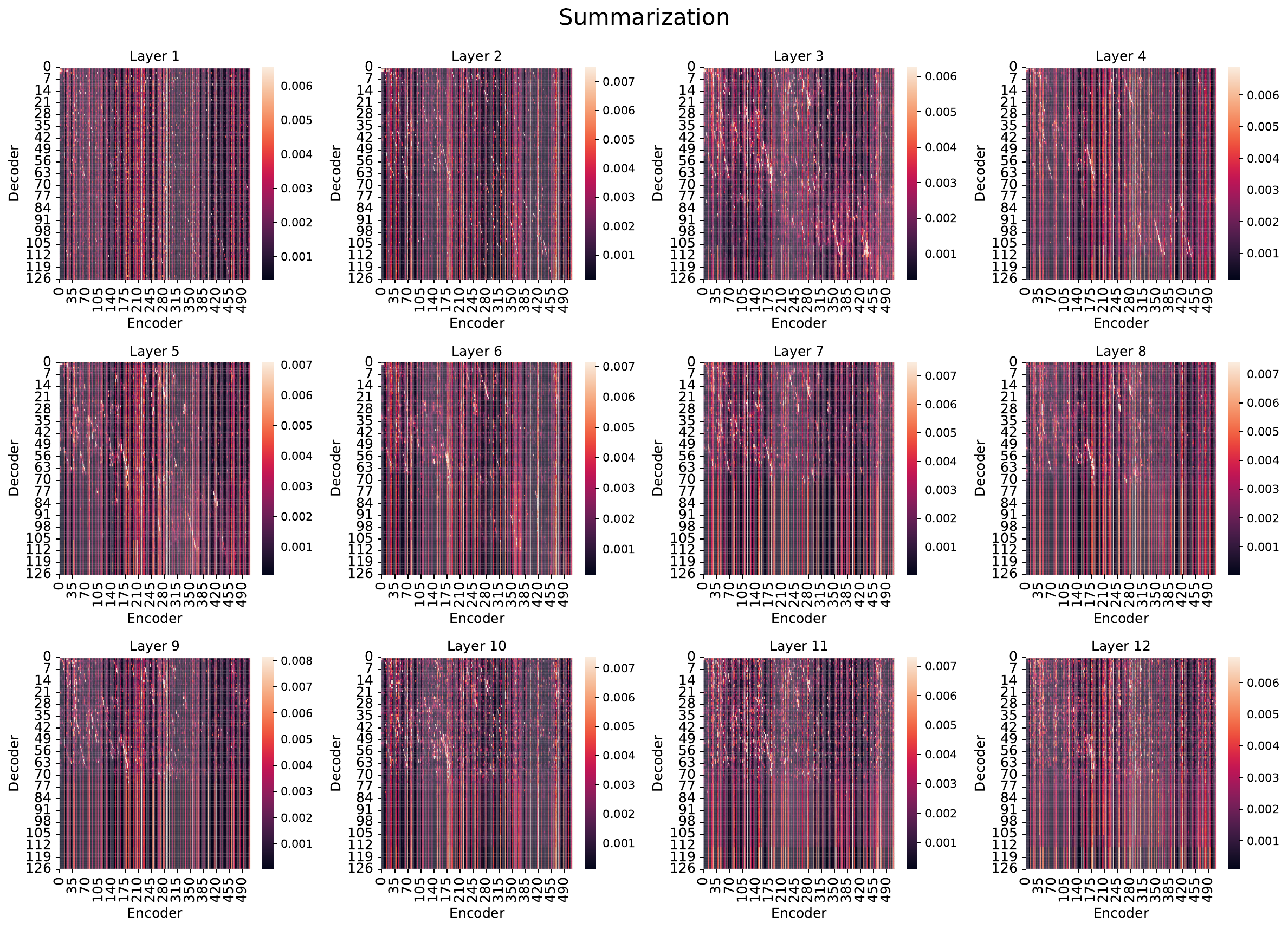}
    
}
\caption{Average decode-encoder attention heatmaps on summarization from different layers}
\label{fig:heatmap-layer-summarization}
\end{figure*}
\begin{figure*}[t]
\centering
\resizebox{\textwidth}{!}{%
    \includegraphics[]{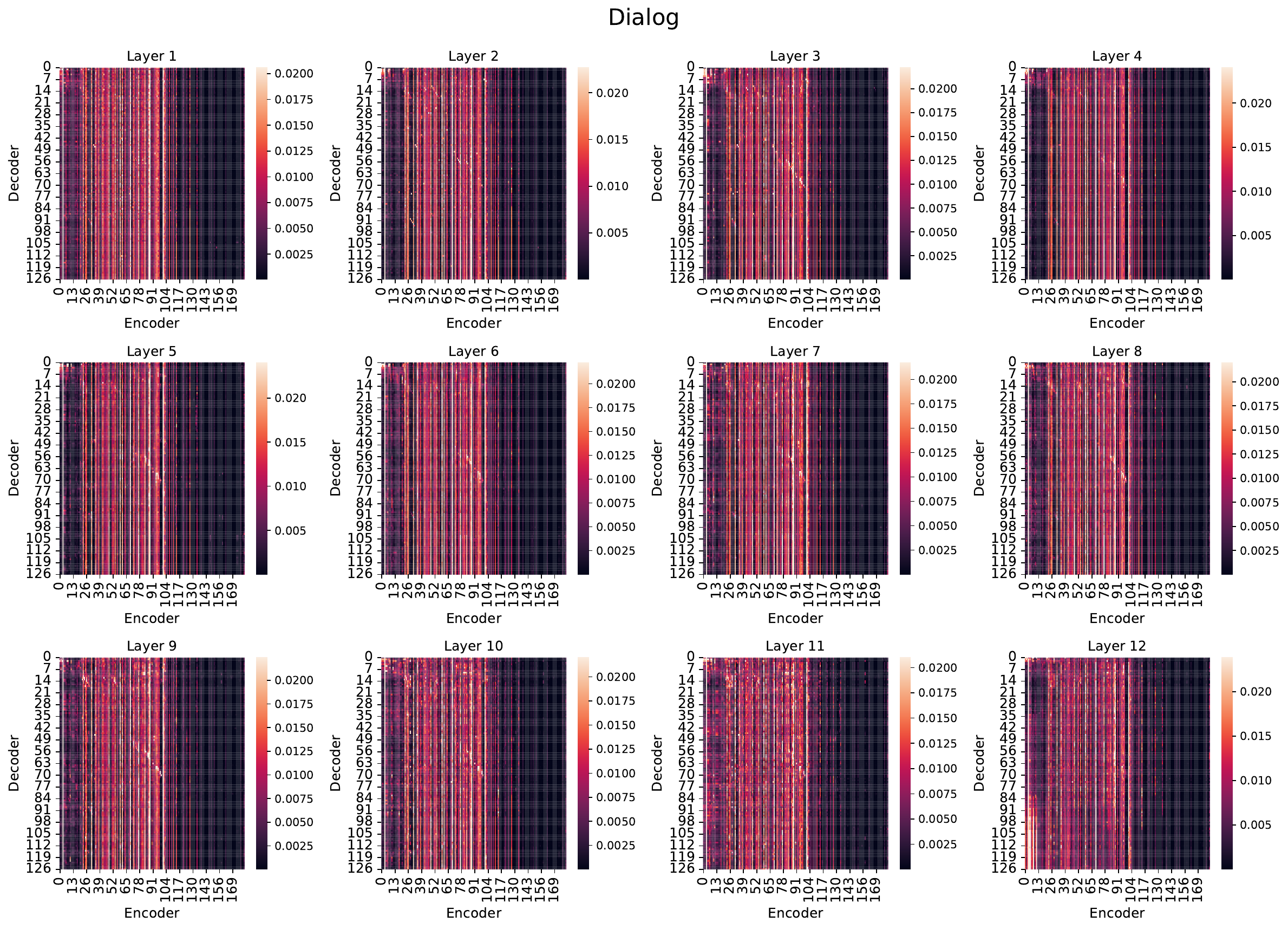}
    
}
\caption{{Average Decode-encoder attention heatmaps on dialog from different layers}}
\label{fig:heatmap-layer-dialog}
\end{figure*}
\begin{figure*}[t]
\centering
\resizebox{\textwidth}{!}{%
    \includegraphics[]{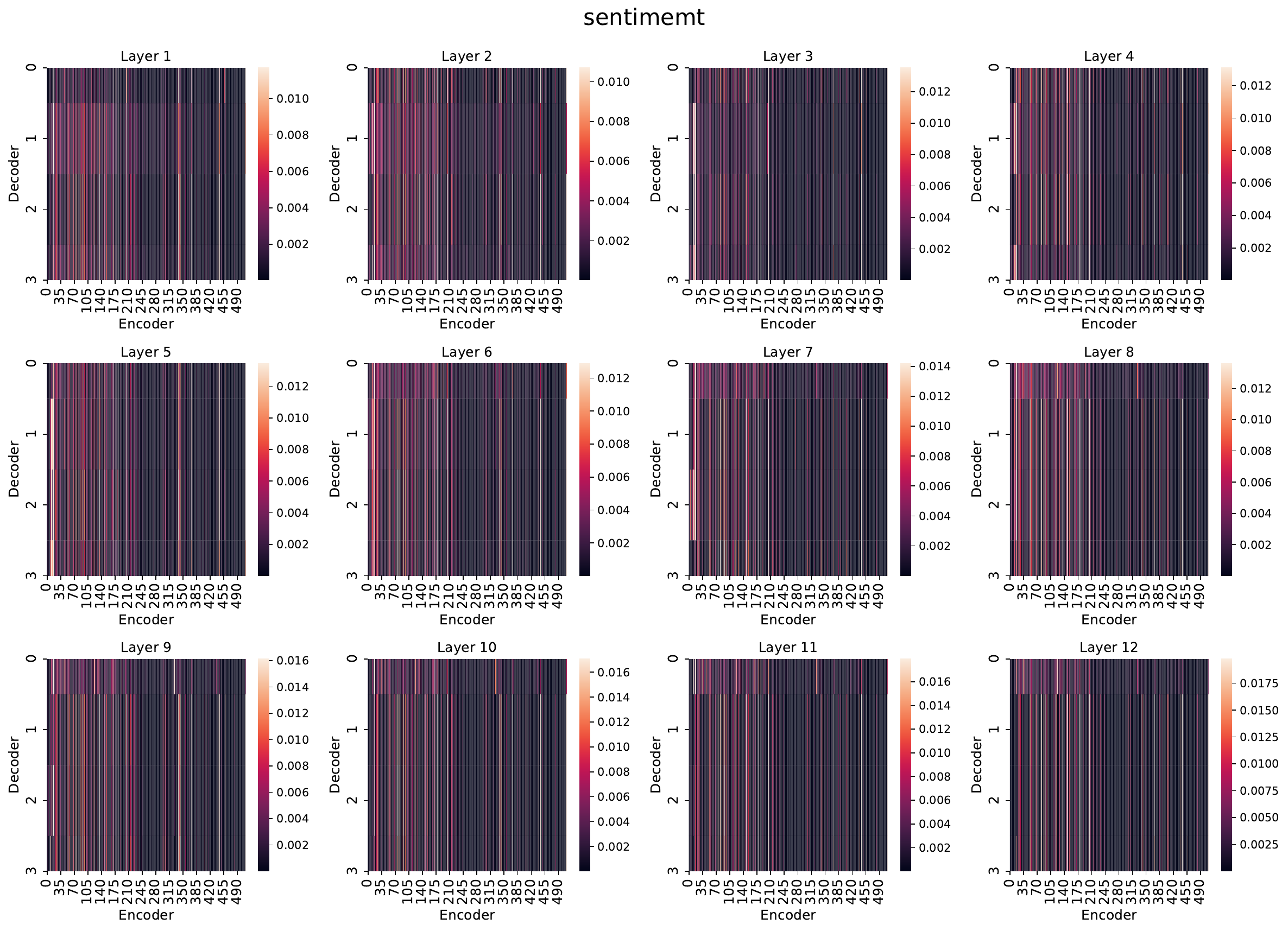}
    
}
\caption{{Average decode-encoder attention heatmaps on sentiment classification from different layers}}
\label{fig:heatmap-layer-sentiment}
\end{figure*}
\begin{figure*}[t]
\centering
\resizebox{\textwidth}{!}{%
    \includegraphics[]{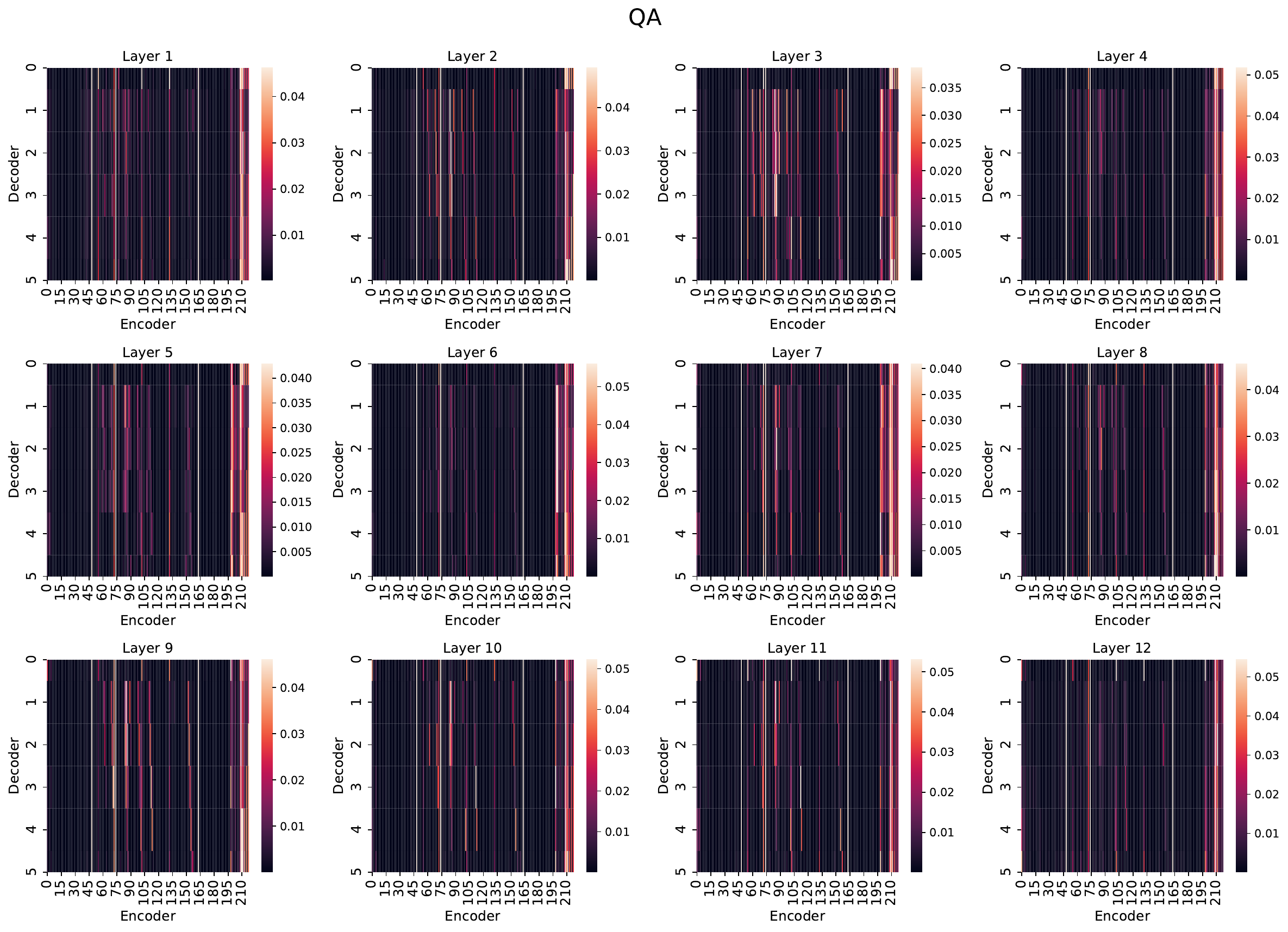}
    
}
\caption{{Average decode-encoder attention heatmaps on QA from different layers}}
\label{fig:heatmap-layer-QA}
\end{figure*}

\label{sec:appendix}

\end{document}